%% file: arxiv.tex
\documentclass{article}

\usepackage[verbose=true,letterpaper]{geometry}
  \newgeometry{
    textheight=9in,
    textwidth=6.75in,
    top=1in,
    headheight=12pt,
    headsep=25pt,
    footskip=30pt
  }

\RequirePackage{times}
\RequirePackage{fancyhdr}
\RequirePackage{xcolor} 
\RequirePackage{algorithm}
\RequirePackage{algorithmic}
\RequirePackage{natbib}
\RequirePackage{eso-pic} 
\RequirePackage{forloop}
\RequirePackage{url}

\usepackage{microtype}
\usepackage{graphicx}
\usepackage{subfigure}
\usepackage{booktabs} 

\usepackage{hyperref}

\usepackage{amsmath}
\usepackage{amssymb}
\usepackage{mathtools}
\usepackage{amsthm}

\usepackage[capitalize,noabbrev]{cleveref}

\usepackage[utf8]{inputenc}
\usepackage[T1]{fontenc}
\usepackage{amsfonts}
\usepackage{nicefrac}
\usepackage{xcolor}
\usepackage{amsmath}
\usepackage{MarkMathCmds}
\usepackage{caption}
\usepackage{enumitem}
\usepackage{algorithm}
\usepackage{multirow}
\usepackage{siunitx}

\theoremstyle{plain}

\theoremstyle{definition}

\theoremstyle{remark}

\newtheorem{recom}{Recommendation}
\newcommand{\thrmspace}{\vspace{1em}}

\newcommand\blfootnote[1]{%
  \begingroup
  \renewcommand\thefootnote{}\footnote{#1}%
  \addtocounter{footnote}{-1}%
  \endgroup
}

\title{Recommendations for Baselines and Benchmarking \\ Approximate Gaussian Processes}

\author{
\normalsize \textbf{Sebastian W. Ober\footnote{Equal contribution.} $^1$\blfootnote{Correspondence to: Mark van der Wilk, \texttt{mark.vdwilk@cs.ox.ac.uk}} 
\quad Artem Artemev$^{*2}$ \quad Marcel Wagenl\"{a}nder$^{*2}$} \\
\normalsize \textbf{Rudolfs Grobins$^3$} \quad \textbf{Mark van der Wilk$^4$} \\
\normalsize $^1$Work done at the University of Cambridge and Secondmind\quad$^2$Imperial College London\\
\normalsize $^3$Work done at Imperial College London\quad$^4$ University of Oxford
}

\begin{document}

\maketitle

\begin{abstract}
Gaussian processes (GPs) are a mature and widely-used component of the ML toolbox.
One of their desirable qualities is automatic hyperparameter selection, which allows for training without user intervention.
However, in many realistic settings, approximations are typically needed, which typically do require tuning. 
We argue that this requirement for tuning complicates evaluation, which has led to a lack of a clear recommendations on which method should be used in which situation. 
To address this, we make recommendations for comparing GP approximations based on a specification of what a user should expect from a method. 
In addition, we develop a training procedure for the variational method of \citet{titsias2009variational} that leaves no choices to the user, and show that this is a strong baseline that meets our specification. 
We conclude that benchmarking according to our suggestions gives a clearer view of the current state of the field, and uncovers problems that are still open that future papers should address.

\end{abstract}

\input{section/intro}
\input{section/goals}
\input{section/near-exact-approximations}
\input{section/baseline-procedure}
\input{section/evaluation}

\input{section/conclusion}

\section*{Acknowledgments}
We would like to thank David R. Burt for helpful discussions.
SWO acknowledges the Gates Cambridge Trust for funding his doctoral studies.

\bibliography{gpbench}
\bibliographystyle{plainnat}

\input{section/appendix}

\end{document}

%% file: section/intro.tex
\section{Introduction}

Gaussian process \citep[GP;][]{rasmussen2006gaussian} models are popular due to their flexibility, interpretability, and powerful uncertainty estimates.
One of their key qualities in regression is their possibility for automatic hyperparameter tuning using the marginal likelihood, which has a closed form.
This quality, combined with powerful quasi-Newton optimizers such as L-BFGS \citep{liu1989limited}, allows for a fully-automated training procedure with essentially no tuning required by the end user.
However, exact GPs require $O(N^3)$ computation and $O(N^2)$ memory, where $N$ is the number of datapoints, making exact computation prohibitively expensive for large datasets.

To address this issue, many solutions have been proposed to make Gaussian processes more scalable.
Of these, sparse ``inducing variable'' approximations \citep{seegerfast,snelson_sparse_2006,quin2005unifying} have remained popular, with the sparse variational methods of \citet{titsias2009variational, hensman2013gaussian, hensman2015scalable} being used as baselines in many papers that propose new approximations. 
This popularity likely stems from the variational method's universal applicability: wherever a GP can be applied, so can a sparse variational approximation.
As a consequence, many papers have been published that claim to outperform sparse GPs. 
This may lead one to wonder why they are still the most common baseline: is there not a better approximation that should be used instead?

The literature does not provide a clear answer to this question. 
Papers are difficult to compare, as they often impose different constraints on methods, leading to papers reporting different results for the same method.
The lack of a clear answer has serious consequences for the GP community: first, it complicates the communication of research to practitioners, and second, it slows down progress by making it difficult for researchers to build on earlier improvements.

We discuss why it is difficult to compare Gaussian process approximations, and how comparisons can be improved. 
To start, we review desired behaviour of GP models and their approximations, in order to find clear ways to perform comparisons (\cref{sec:goals}). 
Based on this, we
\begin{enumerate}[label=\arabic*)]
    \item investigate when GP approximations are close to exact, and how the existence of these regimes should change empirical evaluation (\cref{sec:near-exact});
    \item develop a recommended training procedure for the sparse Gaussian process regression (SGPR) method of \citet{titsias2009variational} that ensures it is a strong baseline which often provides near-exact solutions (\cref{sec:baseline});
    \item develop simple modifications to SGPR to allow for robust training with minimal numerical instability, allowing the user to train SGPR models out-of-the-box without fine-tuning (\cref{sec:sgpr_mods}); and
    \item suggest experimental procedures for two settings of interest which will make it easier to compare evaluations across papers.
\end{enumerate}

The key observation that underpins our recommendations is that good GP approximations should be exact in some computational limit, meaning that they should perform equally well in that limit since they approximate the same GP. 
Differences between methods therefore should only arise from computational bottlenecks, making it crucial for benchmarking procedures to carefully control them. 
Overall, we hope that our recommendations will lead to a clearer demonstration of the strengths and weaknesses of GP approximations within two relevant use cases. 
While researchers may want to consider other use cases as well, we hope to provide a starting point that will increase commonality between papers. 
Such increased clarity is what is needed for users looking for actionable advice on choosing a method.

%% file: section/goals.tex
\section{Goals of Gaussian Process Approximations}
\label{sec:goals}

We consider Gaussian process regression (GPR), where we have observed a dataset containing $N$ observations $(X, \vy)=\{(x_n, y_n)\}_{n=1}^N$ with $x_n \in \Xspace$, an arbitrary input space, and $y_n \in \Reals$. 
We assume $y_n = f(x_n)+ \epsilon_n$, where the $\epsilon_n$ are independent and identically distributed Gaussian random variables. 
Additionally, we take a GP prior over $f$, and write $f|\vtheta \sim \GP(0, k_{\vtheta})$, where $k_{\vtheta}: \Xspace\times \Xspace \to \Reals$ is a covariance (or kernel) function with hyperparameters $\vtheta$. 

Gaussian processes have two main benefits. 
First, they have the ability to automatically select hyperparameters using only a training set, without needing to repeatedly retrain using a validation set. 
This is typically done through type II maximum likelihood using the log marginal likelihood \citep[LML;][Chapter 5]{rasmussen2006gaussian}:\footnote{In extreme cases maximum marginal likelihood will overfit \citep[see for example][]{oberpromises21}; however, this is not generally the case for Gaussian process regression with commonly used kernels and large numbers of observations.}
\begin{align}
    \vtheta_{\text{opt}} = \argmax_{\vtheta} \log p(\vy|X, \vtheta).
\end{align}
Second, GPs have good uncertainty quantification properties, through a combination of Bayesian inference and their non-parametric nature, i.e.,~their equivalence to an infinite basis function model \citep{rasmussen2006gaussian}. We take advantage of this by predicting with the posterior
\begin{align}
    p(f(X^*)|\vy, X, \vtheta_{\text{opt}}) \,.
\end{align}
Both the posterior and the marginal likelihood have closed-form Gaussian densities, which to evaluate require decomposing an $N\!\times\! N$ kernel matrix, $K_{XX} + \sigma^2 I_N$. 
This gives exact implementations of GPR regression an $O(N^3)$ computational cost and an $O(N^2)$ memory cost.
We give a more in-depth treatment of GPR in \cref{app:gpr}.

\subsection{Sparse Gaussian Process Regression (SGPR)}\label{sec:sgpr-properties}

As these costs are often prohibitive for larger datasets, approximations with lower computational and/or memory cost are often used.\footnote{In this paper, as in most prior work, we focus primarily on the computational/time cost of GPs. However, in practice, lack of memory may be a larger obstacle than computational time, especially as out-of-memory errors will halt computation immediately. 
Implementations of SGPR that are intrinsically more memory efficient have long existed \citep{gal2014distributed} but were cumbersome, but recent tools can provide these benefits without modifying code \citep{artemev2022memory}.
Alternatively, running on CPU may provide the user more memory, but at the cost of a longer run time.}
In this work, we place particular emphasis on the sparse Gaussian process regression \citep[SGPR;][]{titsias2009variational} approximation, which we use for our proposed baseline, and which we briefly summarise here (for more details, we refer the reader to \cref{app:sgpr}).

SGPR is a variational approximation \citep{blei2017variational} that relies on $M$ inducing points, with $M \!\leq\! N$, defined at inducing locations $Z \!=\! \{z_m\}_{m=1}^M$.
These inducing points attempt to summarise the full dataset, allowing inference to be performed by decomposing a smaller $M \!\times\! M$ kernel matrix $K_{ZZ}$. 
As SGPR is a variational approximation, it provides a lower bound (known as the ELBO) to the LML (\cref{eq:sgpr_elbo}) and an approximate predictive posterior (\cref{eq:sgpr-pred}). 

The quality of the approximation is therefore controlled by two parameters: the number of inducing points $M$, and the inducing point locations $Z$. 
The computational cost is controlled by $M$, giving $O(NM^2 + M^3)$ time complexity, and $O(NM + M^2)$ memory cost. 
The degree of speed-up therefore depends on how large $M$ needs to be in order to get a good approximation. 
Theoretical analysis has shown that as $N\to\infty$, we can have $M\ll N$ while still obtaining an arbitrarily exact approximation \citep{burt2019rates,burt2020convergence}, but such asymptotic guarantees do not directly help when dealing with finite datasets.
Fortunately, SGPR has the following helpful properties:
\begin{enumerate}[label=\arabic*)]
    \item The ELBO is a lower bound to the true LML, and its quality monotonically increases as $M$ increases \citep{matthewsthesis2016,bauer_understanding_2016}.
    \item As a variational method, the difference between the true LML and the ELBO is equal to the KL divergence of the approximation $Q$ to the true posterior $\tilde P$, $\log p(\vy|X,\vtheta) - \text{ELBO} = \mathrm{KL}[Q||\tilde{P}]$. This means that this quality measure of the posterior also monotonically improves with $M$ as measured by the KL.
    \item A near-exact solution will be achieved once $M$ is large enough, since the true posterior is recovered for $M=N$ and $Z=X$.
    \item An upper bound on the LML can be computed in the same time and memory costs as the ELBO \citep{titsias_variational_2014}, giving an upper bound on the KL that is useful in real settings (\cref{eq:sgpr-upperbound}).
\end{enumerate}
In Section~\ref{sec:baseline}, we use these qualities of SGPR to recommend a training procedure which ensures that it is a strong baseline method which does not require tuning by the user.

Having summarised the necessary technical background for this work, we now turn to describing what a Gaussian process approximation should achieve.

\subsection{Desiderata for GP Approximations}\label{sec:gp-desiderata}
As described above, exact GPR allows us to automatically select hyperparameters without the need of a validation set, while maintaining useful uncertainty estimates.
A good approximation should therefore aim to provide both these benefits of exact GPs while reducing the computational and memory costs and adding as few complications (e.g.,~tunable parameters) as possible. 
Based on this, we propose the following desiderata that GP approximations should satisfy:
\begin{enumerate}[label=\arabic*)]
    \item An method for automatic hyperparameter selection. In many cases, this will be achieved through an approximation to the (log) marginal likelihood.
    \item Accurate approximation of the predictions (mean and variance) at these hyperparameters, using as little computational resources as possible.
    \item A transparent experience for the user, by requiring as little adjustment as possible.
\end{enumerate}

Currently, it is common to assess hyperparameter learning and prediction jointly by evaluating on predictive metrics only. 
Most commonly, root-mean-square error (RMSE) and negative log predictive density (NLPD) are used, where RMSE is used to assess pure predictive performance (i.e., the mean prediction), with NLPD being used to assess the uncertainty quantification in conjunction with the predicted mean. 
The computational cost in these assessments is typically controlled by an approximation-dependent parameter, for instance the number of inducing points for variational methods, or conjugate gradient iterations for iterative methods \citep[e.g.,][]{wang2019exact}. 
The results are typically presented in a table that shows the final predictive performance and runtime of different methods.
This approach to evaluation is justified by the fact that GPs are used to make predictions, and so an approximation that is superior in terms of predictions is all that is needed. 

In the following, we argue that this approach is incomplete for ensuring the desiderata are met, and that authors should
\begin{enumerate}[label=\arabic*)]
\item be aware that improved predictive performance does not imply a better Gaussian process approximation \citep[as is known for e.g., the FITC approximation][]{snelson_sparse_2006,bauer_understanding_2016}, and
\item take care to benchmark methods across a range of computational budgets, to illustrate their efficient frontiers.
\end{enumerate}
We develop recommendations that will allow all three desiderata for GP approximations to be clearly assessed.

\subsection{Predictive Approximation Quality}
Given our requirements, an empirical evaluation should address the question: ``Does the approximation provide solutions that are close to exact GPR?''
The typical evaluation procedure does not directly answer this question, as good predictions on held-out data do not necessarily indicate that the approximation is close to exact GPR. 
Indeed, in some cases an approximation can outperform the original model in terms of predictive metrics.
This behaviour can come from a difference in the hyperparameter selection procedure, or in a difference in the predictive approximation, or both.

If the approximation provides different predictions to the true model \emph{for the same hyperparameters}, then the approximation's accuracy to the true model is limited, regardless of how well the approximation performs.
Usually, improved performance for the same hyperparameters occurs if the approximation has more freedom to fit the training data than the original GP model, which may be the case particularly when the model is misspecified. 
The FITC approximation \citep{snelson_sparse_2006} is a well-known example of this behaviour, as the approximation can fit heteroskedastic noise, even though the original GP cannot. 
This effect happens because the approximation conflates uncertainty in the function with noise \citep{bauer_understanding_2016}, meaning that while it can give better predictions in terms of test metrics, it is a poor approximation to the original GP.

It is debatable whether such behaviour is desirable. 
On the one hand, this behaviour can be seen as fixing a problem with the original model, which can be a contribution in its own right. 
On the other hand, such behaviour is an example of an inaccurate approximation, which can also cause unexpected behaviour in other settings. 
For example, if an approximation is more free to fit the data than the original GP, it may also be more susceptible to overfitting. 
Either way, these effects should be made clear to the user, and explicitly investigated and discussed.

\thrmspace
\begin{recom}
\label{re:gp-comparison}
Assessments of an approximation should report metrics indicating how close it is to the exact solution, when possible.
This is important to quantify the fidelity of the approximation.
\end{recom}
\thrmspace
This can be achieved by computing (bounds on) distances to the exact solutions, when computationally tractable.
Past work has considered KL divergences to exact predictions \citep{titsias_variational_2014, kim_scaling_2018,burt2022thesis}, but other measures may also be appropriate, for instance, Wasserstein distances \citep{mallasto2017wasserstein,wilson2020matheron}; the appropriate choice of metric depends on the particular properties of the posterior the user wants to preserve \citep[see][for a related discussion]{huggins2020validated}.
Additionally, it can be useful to evaluate on toy datasets that are designed to highlight a specific behaviour. 

\subsection{Assessing Hyperparameter Selection}
Differences caused by hyperparameter selection need to be disentangled further, which is complicated by the fact that hyperparameter optimisation is non-convex. 
This means that the solution found by an exact GP with gradient-based optimisation may not actually be the global solution. 
It is possible for approximations to both find hyperparameters that would be better \emph{or} worse optima for the exact model \citep[\S 3.5]{bauer_understanding_2016}. 
Moreover, it is also possible for model misspecification to allow an approximation to select hyperparameters which give better predictive metrics, but which the true GP would not select.
These three scenarios should be distinguished, which can be achieved by again following \cref{re:gp-comparison}. 

\subsection{Evaluating Approximate GPs}
While predictive metrics alone are not sufficient for determining the quality of an approximation, investigating predictive metrics is still necessary, since 1) predictive metrics are typically the most relevant to solving a task; 2) poor approximations often lead to poor predictions; and 3) it may be difficult to obtain exact GP predictions for comparison. 
However, it is also not straightforward to perform comparisons across methods, as approximations often introduce additional parameters which control the trade-off between computation and accuracy. 
Typically, different datasets require different parameter values for the best performance, which raises questions about the extent to which competing methods should be tuned, and how the cost of tuning should be included in an evaluation. 
Needing to tune methods is also incompatible with our requirement that approximations be transparent to the user.
One solution that provides both a transparent experience to the user and fair comparisons is to require methods to provide an automatic procedure for selecting approximation parameters.\footnote{\citet{rasmussen1997thesis} also discussed the importance of automatic methods for setting parameters in benchmarking.}
\thrmspace
\begin{recom}
\label{re:approxparams}
A method should contain a well-defined procedure for setting any approximation parameters. 
This should be assessed and benchmarked as an integral part of the method.
\end{recom}
\thrmspace

For instance, cross-validation can always be applied when approximation parameters need to be tuned. 
However, using a validation set to tune parameters typically requires retraining the model repeatedly with different settings. 
This repetition can greatly increase the cost of using the method, and must be taken into account when comparing to other approaches. 
Alternatively, default values can be used, and are successful if a single setting generally leads to good performance across many situations (i.e.,~datasets and models). 
For some approximation parameters (e.g.,~number of inducing variables or CG iterations), it is known that increasing them monotonically improves the approximation quality. 
These can be tuned by continuously increasing them as more computation is provided.

When this is the case, the approximation method allows users to trade off computation and accuracy. 
Such methods are therefore best compared by considering the Pareto frontier determined by evaluating the method using different amounts of compute. 
This allows a practitioner to consider the marginal cost of obtaining slightly more accuracy in their approximation, enabling them to take their own goals and constraints into consideration.
Two settings are often relevant: 1) a computation-constrained practitioner wants an answer given a particular compute budget; and 2) a practitioner wants to get within some distance of the optimal performance, and is willing to wait as long as it takes to do so.
This informs our next recommendation:

\thrmspace
\begin{recom}
\label{re:eval-settings}
Approximation methods should be evaluated by \textbf{1)} measuring the performance for various compute budgets; and \textbf{2)} measuring the compute needed to achieve a particular performance goal, where every effort is taken to provide enough compute.
\end{recom}
\thrmspace

%% file: section/near-exact-approximations.tex
\begin{figure*}[t]
    \centering
    \includegraphics[width=\textwidth]{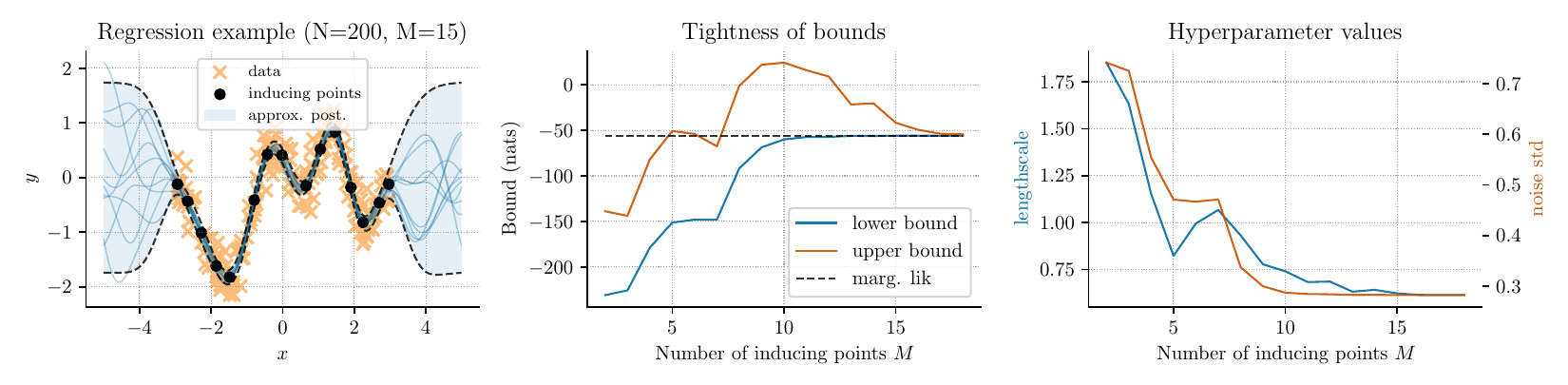}
    \caption{SGPR with a squared exponential kernel (highest true marginal likelihood) on the toy 1D \texttt{Snelson} dataset. \textbf{Left}: Example approximate solution. \textbf{Middle}: Upper and lower bounds on marginal likelihood with varying $M$. Note that different hyperparameters are found as $M$ increases, which allows the upper bound to rise, before it eventually converges as the hyperparameters converge. \textbf{Right}: Hyperparameters with varying $M$.}
    \label{fig:sgpr-working-toy}
\end{figure*}
\section{Near-Exact Approximations}
\label{sec:near-exact}

The typical situation for a Bayesian model is that inference is \emph{analytically} intractable. 
Approximate inference schemes are introduced to find some tractable distribution (e.g.,~a Gaussian) to approximate a posterior that has no manageable closed-form solution. 
Gaussian process regression provides a unique scenario for approximate inference, as the true posterior \emph{is analytically tractable} (Gaussian), but \emph{computationally expensive}. 
As a consequence, GP approximations can often be arbitrarily accurate.

For example, conjugate gradient (CG) methods \citep{gibbs1997efficient,davies2015thesis,wang2019exact} converge to the exact solution when given sufficient iterations, SGPR with enough inducing points \citep{burt2019rates,burt2020convergence}, interpolation methods \citep{wilson2015kernel} with a dense enough grid, and random Fourier feature-based methods \citep{rahimi2007rff,lazaro-gredilla2010ssgp} if sufficient features are used. 
One commonality between all these results is that ``enough'' computational resources must be added. 
What is ``enough'' is dataset-dependent, and impractical to predict beforehand.\footnote{While \emph{a priori} results are known for some methods ~\citep[e.g.,][]{burt2020convergence}, they require detailed knowledge of the data generating process, and involve impractically large constants to provide practical recommendations.} 
The importance of recovering the true posterior in practice has been discussed before \citep[e.g.,][]{wilson2015thoughts,bauer_understanding_2016,matthewsthesis2016,vdw2019thesis}, but, we argue, should be given more importance in empirical evaluation.

\subsection{Datasets Where Near-Exactness is Achievable}
Here, we investigate whether there are datasets where a ``near-exact'' approximation can be made while still retaining significant computational and memory benefits over exact GPR.
\Cref{fig:sgpr-working-toy} illustrates an example of this behaviour for an $N=200$ toy 1D dataset for SGPR. 
For $M=12$ we already see convergence of the ELBO, which is necessary for near-exactness. 
With a few more inducing points ($M>16$), we see the upper bound converges as well, proving that the predictive posterior is near-exact for these hyperparameters. 
By comparing to an exact GP implementation, we also verify that we haven't converged to a different (or worse) local optimum, following \cref{re:gp-comparison}. 

This example shows that near-exact approximations are possible.
Therefore, all approximations that achieve this for a given dataset should give the same results, resulting in a useful consistency check between approximations. 
In addition, showing that a new approximation can produce similar results is evidence for its usefulness. 
However, this regime also limits the usefulness in comparing measures of predictive performance between methods, since they perform equivalently.
Indeed, the main takeaway from comparing two near-exact approximations is in how efficient they are with respect to computational resources and memory usage.
\thrmspace
\begin{recom}
\label{re:near-exact-comparison}
Approximations should be compared by observing for how many datasets near-exactness can be reached, and how much compute is required to find such a solution in the near-exact case.
\end{recom}
\thrmspace
If two near-exact posterior approximations yield different results, it may be interesting to investigate the reason for the improvement: comparing multiple near-exact methods allows the source to be identified. 
For example, if two posteriors are near-exact, one can investigate whether hyperparameter optimisation is the cause of differing predictions by transferring hyperparameters from one method to another.

\begin{figure*}[t]
     \centering
    \includegraphics[width=\textwidth]{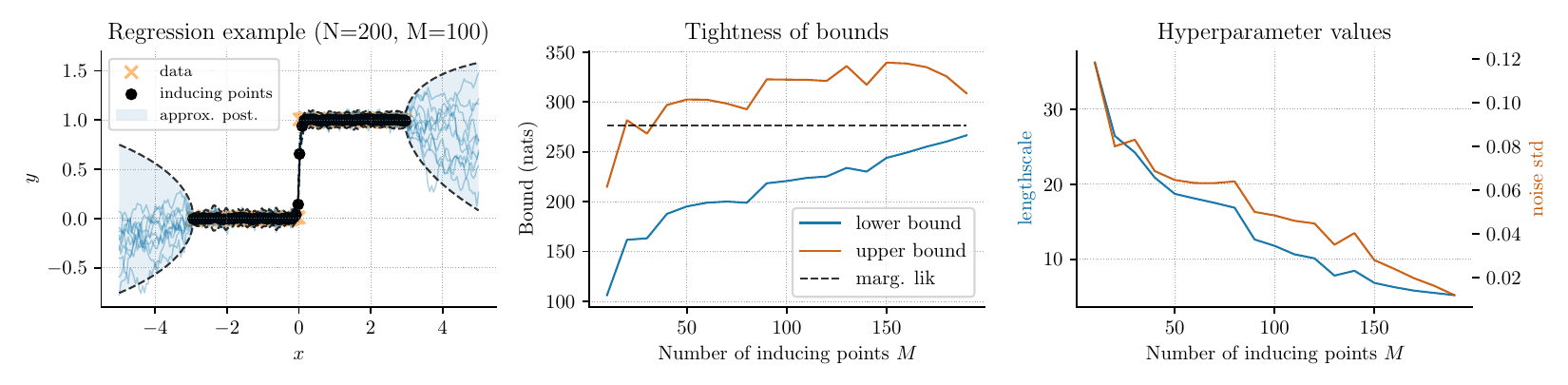}
    \caption{SGPR with a Mat\'{e}rn-$\frac{1}{2}$ kernel (highest true marginal likelihood of stationary kernels) on a step dataset. \textbf{Left}: Example approximate solution. \textbf{Middle}: Upper and lower bounds on marginal likelihood with varying $M$. Note that different hyperparameters are found as $M$ increases. The upper and lower bounds do not converge. \textbf{Right}: Hyperparameters with varying $M$, which do not converge even when $M \approx N$.}
    \label{fig:sgpr-failing-toy}
\end{figure*}

\subsection{Datasets with no Near-Exact Approximation}
\label{sec:sgpr-fails}
SGPR does not behave as well on all datasets, particularly when the full kernel matrix is not low-rank at the optimal hyperparameter setting. 
This often occurs when data is sparse, or if the model is misspecified. 
An example of the latter can be seen in \cref{fig:sgpr-failing-toy}, where neither the ELBO nor the upper bound converge, even for large $M$. 
In this example, model misspecification causes the lengthscale to continuously shrink, which makes the kernel less low-rank.\footnote{If misspecification occurs, then perhaps a better kernel should be sought, rather than a better approximation. We provide an illustration of a better kernel for this dataset in \cref{fig:sgpr-working-relu}.}

Additional insight can be provided by investigating when an approximation provides near-exact solutions, and when it does not.
This helps to identify different regimes where approximations are appropriate: for instance, iterative conjugate gradient methods can often efficiently solve methods where sparse methods would fail \citep{burt2022thesis}.
Moreover, comparing methods on how many near-exact solutions they provide helps remove selection bias of the datasets that are benchmarked on.

%% file: section/baseline-procedure.tex
\section{SGPR as a Strong Baseline}
\label{sec:baseline}

Taking into account our recommendations, we now turn to developing a training procedure for SGPR \citep{titsias2009variational} that makes it suitable as a robust baseline method. 
Compared to how it is commonly applied, some small tweaks significantly strengthen it, and allow more insight to be gained from experiments.
For baselines, this is particularly important, since many papers rely on automatically running methods on many datasets to demonstrate success.

SGPR is typically trained by selecting a value for $M$, after which the variational parameters $Z$ and the hyperparameters $\vtheta$ are trained together to maximise the ELBO. 
This cannot be expected to work universally well, as different datasets with different sizes and properties will require different values of $M$, as illustrated above. 
To address this, our main suggestion is to continuously increase $M$ throughout training. 
This allows us to satisfy the both parts of \cref{re:eval-settings}: for the first part, by allowing results to be obtained with an increasing computational budget. 
For the second part of \cref{re:eval-settings}, we would like to ensure that the method converges to the exact solution if $M$ grows large enough. 
The first two properties (cf. Sec.~\ref{sec:sgpr-properties}) of SGPR indicate that this is possible, as long as $Z$ is chosen well enough \citep{burt2020convergence}. 

To achieve this, our procedure starts by selecting $Z$ using the greedy variance technique of \citet{burt2020convergence}. 
This procedure is related to determinantal point process sampling, which ensures fast convergence to the true posterior \citep{burt2019rates}, and also provides higher ELBO values at initialisation than alternatives such as k-means or uniform subsampling \citep{burt2020convergence}. 
Following this initialisation, we maximise the ELBO with respect to the hyperparameters $\vtheta$, using the parameter-free quasi-Newton optimiser L-BFGS \citep[\S 7.2][]{nocedal2006numerical,virtanen2020scipy}.
Following the procedure proposed by \citep{burt2020convergence}, we then repeat this process of initialising the inducing locations (with the new hyperparameters) followed by optimising only the hyperparameters, until convergence is reached. 
This avoids the increased complexity of joint optimisation of inducing locations and hyperparameters.
We note that the inputs for our procedure are the same as for an exact GP implementation, thereby satisfying \cref{re:approxparams}. 
However, a couple of numerical issues remain that prevent the universal application of this procedure.

\begin{figure*}[t]
  \centering
  \includegraphics[width=\textwidth]{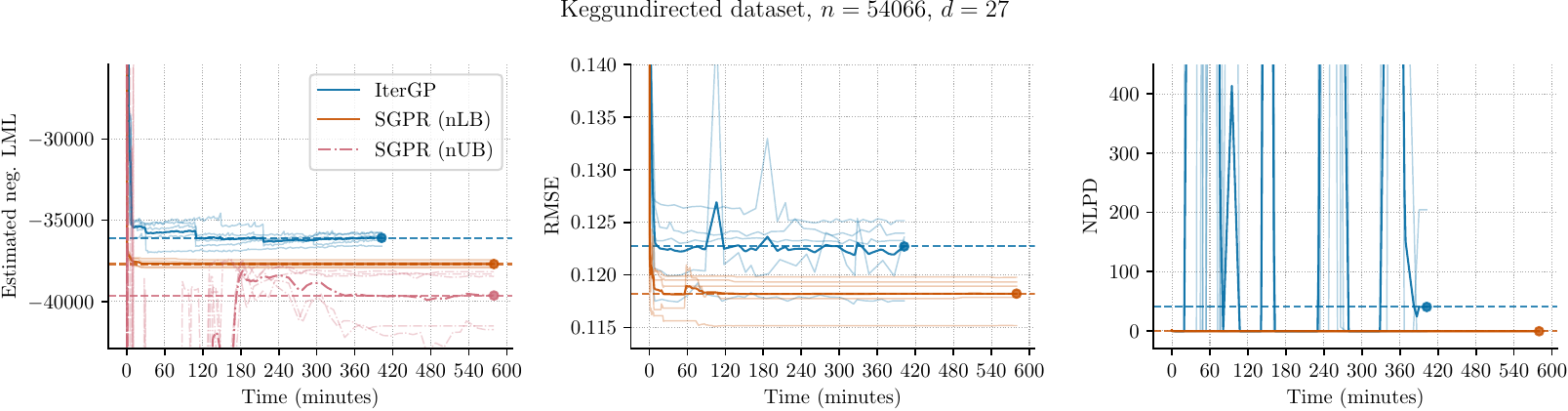}
  \caption{Illustration of our proposed benchmarking procedure on the \texttt{Keggundirected} dataset, where SGPR yields a near-exact approximation. We plot the results from five independent runs for each method, plotting the negative ELBO (nLB), negative upper bound (nUB), negative approximate LML (for IterGP), and RMSEs and NLPDs for both methods. We also provide training set size $n$ and input dimension $d$ for reference. Lower is better for all metrics.}
  \label{fig:near-exact-UCI}
\end{figure*}

\subsection{Improving SGPR's Numerical Stability}
\label{sec:sgpr_mods}
Perhaps the most prevalent obstacle to the fully automatic use of SGPR is that of numerical stability.
Numerical instability typically arises due to the need to invert the Gram matrix $K_{ZZ}$ implied by the GP kernel without added noise, which can often fail when $M$ is large or when the inducing locations are close to each other.
To resolve this, a small amount of ``jitter'' is often introduced to stabilise the computation of the inverse, so that we invert $K_{ZZ} + \epsilon I_M$ instead, where $\epsilon$ is small.
However, we typically want to minimise the amount of jitter, as jitter amounts to adding noise to the data and harms the ELBO \citep{titsias2009variational}.
Nevertheless, varying amounts of jitter will be required for different problems and the amount of jitter needed can vary as the hyperparameters change.
Therefore, we follow prior works \citep[see e.g.,][]{gpy2014,gardner2018gpytorch,burt2020convergence} in adaptively increasing the jitter during inversion to ensure that the Gram matrix is numerically positive-definite.

While the use of adaptive jitter mitigates numerical errors, we still found that some numerical errors persisted.
In prior work, this would be mitigated by placing upper bounds on the allowed values of certain hyperparameters \citep[see e.g.,][]{burt2020convergence}.
However, we again found that the most suitable values of these bounds would vary depending on the dataset, inhibiting a truly automatic procedure.
Upon further inspection, we found two added sources of numerical errors.
First, the trace term in the ELBO, $\mathrm{tr}(K_{XX} - Q_{XX})$, which must be greater than or equal to zero mathematically, would occasionally become negative.
We address this by manually setting the term equal to zero when the errors are small.
Secondly, the L-BFGS optimiser, which builds an approximate Hessian from the history of objective function and gradient evaluations, would occasionally suggest extreme hyperparameter values to query, leading to numerical failure.
In this case, we restart the optimiser by clearing the history and resetting the Hessian approximation.
We discuss these changes in further detail in \cref{app:numerical-stability}.

\begin{figure*}[t]
  \centering
  \includegraphics[width=\textwidth]{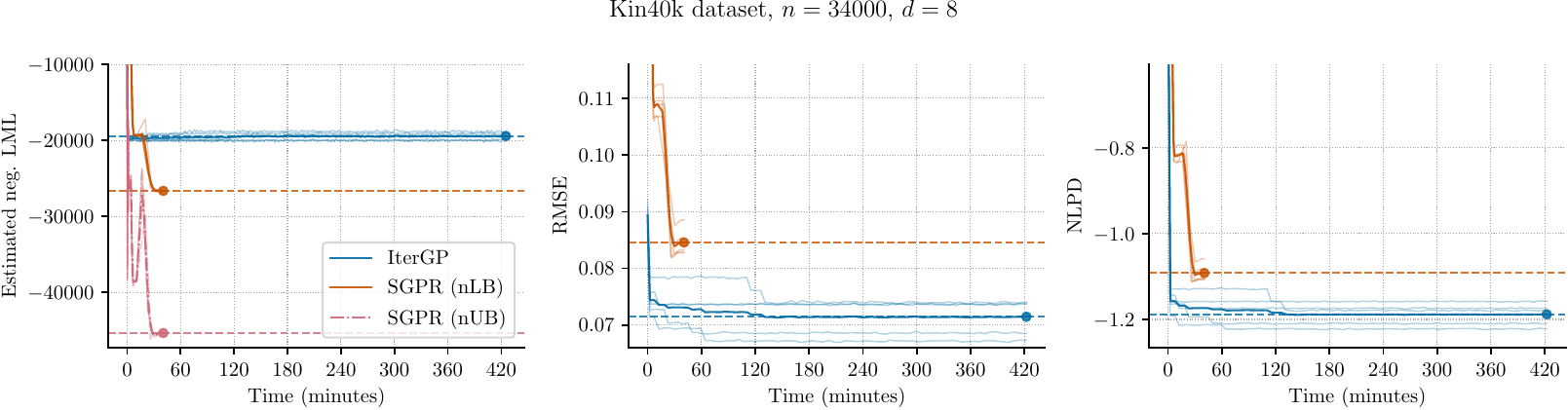}
  \caption{Our proposed benchmarking pocedure on \texttt{Kin40k}, where SGPR does not give a near-exact approximation.}
  \label{fig:non-exact-UCI}
\end{figure*}
\subsection{Minibatch Training of GPs}
Recent advances in sparse variational inference for GPs have allowed for both minibatching \citep{hensman2013gaussian} and non-Gaussian likelihoods \citep{hensman2015scalable}, greatly increasing the applicability of Gaussian processes in terms of dataset size and types of problems.
As such, these stochastic variational GPs (SVGPs) have perhaps become more popular than SGPR as baselines for other methods to compare to, with many works claiming to outperform them \citep[see e.g.,][for recent examples]{lin2024sampling,wu2023largescale}.

Despite this broader applicability, we argue that SGPR is a better baseline for a number of reasons.
First, while SGPR has a higher per-iteration computational cost, it converges in few iterations.
By contrast, SVGP can take far more gradient steps to converge, as the added stochasticity in the ELBO harms variational inference's convergence \citep{wang2023joint}.
Additionally, as SGPR automatically integrates out the optimal variational parameters, SGPR has far fewer parameters to optimise than SVGP, resulting in a simpler optimisation problem.
Indeed, we found that SGPR is typically faster in terms of computational time (see below).
Perhaps more problematic is that SVGP introduces many more tunable hyperparameters, making it difficult to provide a universal recommendation for a large range of datasets (cf. \cref{re:approxparams}), conflicting with our requirement for a transparent user experience (cf. \cref{sec:gp-desiderata}).

While we argue against the use of SVGP as a baseline, we recognise that it remains an important tool in the practitioner's toolbox.
Therefore, we provide some recommendations backed by empirical results for using SVGP models effectively in \cref{app:svgp-exp}.
In short, we recommend that 
\begin{enumerate}[label=\arabic*)]
    \item the minibatch size should be made as large as feasible to reduce minibatching noise;
    \item for convergence, optimisation should be run for as long as feasible (much longer than typically suggested in the literature), using a learning rate scheduler to gradually decrease the learning rate; and
    \item if using the Adam optimiser, improved initial convergence speed can often be achieved by setting the learning rate as large as feasible and the momentum hyperparameters to $\beta_1 = \beta_2 = 0.5$; whether changing the momentum parameters is effective for the entire run depends on the minibatch size and learning rate.
\end{enumerate}
Despite these recommendations, in our experiment in \cref{app:svgp} we find that no setting in our grid search results in a method truly comparable to SGPR in terms of its speed of convergence.
Furthermore, we believe that optimal learning rates and momentum parameters will still vary across datasets, making the task of finding a universal setting that performs well across different settings difficult.
These observations reinforce our belief that SGPR is more well-suited to being a baseline method, whereas SVGP is more suitable as a method in settings where SGPR cannot be applied (for instance, for very large datasets or classification likelihoods).

%% file: section/evaluation.tex
\section{Timed Performance Evaluation}\label{sec:experiments}

To provide an example of how benchmarking for GP approximations should be approached, we run our baseline procedure for a squared exponential kernel GP together with a CG-based iterative GP  (which we term IterGP) approximation \citep{wang2019exact} in a timed performance evaluation on UCI datasets, measuring the performance at multiple time points.
We note that this tests all aspects of the approximations, including their implementations.
Default values for IterGP's two free parameters, CG residual norm tolerance and preconditioner size, suggested in the past have been noted to lead to convergence issues or poor performance \citep{potapczynski21a,artemev2021tighter}. 
We use the parameter settings found by \citet{maddox2021reliablyaccurate}, who tuned the parameters on the UCI datasets to convergent training and good performance.
We run each method over 5 different seeds, and additionally use exact GPR baselines, where possible. and additionally use linear regression, constant function, and, where possible, GPR baselines. 

We give an example of plots showing results for the \texttt{Keggundirected} and \texttt{Kin40k} datasets in \cref{fig:near-exact-UCI,fig:non-exact-UCI}.
We plot LML approximations (including upper bounds for SGPR), RMSEs, and NLPDs.
For \texttt{Keggundirected}, we observe from the ELBO that SGPR quickly gives a near-exact approximation, where the final approximation has $M=10,000$ (although the upper bound did not fully converge for all runs).
Meanwhile, the IterGP gives a similar LML approximation, but worse predictive metrics, with the NLPD showing extremely unstable behaviour.
By contrast, for \texttt{Kin40k}, we observe that SGPR does not have a near-exact solution, whereas the IterGP converges quickly (albeit to a worse LML approximation).
We repeat our benchmarking for 10 other UCI datasets of medium size in \cref{app:fig}, and provide tabulated results in \cref{app:tables}, with the metrics given at sensibly-defined checkpoints to illustrate the methods' time-accuracy trade-offs.
Observing the results from these datasets, we find that SGPR can be near-exact for the following datasets (based off the necessary condition of ELBO convergence): \texttt{Elevators}, \texttt{Keggdirected},  \texttt{Naval}, and \texttt{Skillcraft}.
We also observe the importance of including linear and constant function baselines (\ref{app:tables}): by achieving equivalent performance, these indicate that \texttt{Tamielectric} may not be suitable for SE GPs.

By contrast, even with the improved settings of \citet{maddox2021reliablyaccurate}, we observe that the performance of the IterGP is often erratic, with drastic, unpredictable spikes in many of the metrics.
It seems to do most poorly where SGPR has a near-exact approximation, indicating potential complementarity between the methods.
Conversely, it generally seems to perform best where SGPR does not perform as well, particularly on \texttt{Poletele}, \texttt{Power}, and \texttt{Protein}, albeit with inconsistent approximate LMLs.
For both approximations, we note that there is often not a clear correlation between approximating the true GP and improving predictive performance, reinforcing our original motivation for a clearer benchmarking procedure.
Finally, we note that when feasible, generally the best option remains exact GPR, as it is faster than both approximations (although GPR fails on \texttt{Naval}).

%% file: section/conclusion.tex
\section{Conclusion}

SGPR is a strong baseline, for which there is strong evidence that it achieves near-exact performance for many datasets.
The main reason for its usefulness as a baseline is that no human intervention is needed as its approximation quality is continuously improved: it only needs to be allowed more compute, in the form of inducing points. 
It would be beneficial to develop similar procedures for other approximations. 
As a baseline, SGPR may sometimes still require too many inducing points for it to be practical (\cref{sec:sgpr-fails,fig:sgpr-failing-toy}), which provides opportunities for other methods.

More importantly, we suggest a procedure for comparing GP approximations which provides insight into \textbf{1)} inherent properties of an approximation, such as whether it can be near-exact; and \textbf{2)} what the full time-performance trade-off is, which is actionably useful for users. 
This provides a more complete picture than only reporting results for a single fixed computational budget (e.g.,~fixed inducing points), which is currently common.

Now that GP approximations are becoming very accurate, we believe that such a thorough benchmarking protocol should be standard. 
Methods should be published based on practical strength, \textbf{or} beneficial mathematical properties. 
Our recommendations give a view of both, which we believe would be an improvement over current common standards. 
Nevertheless, there are still important effects that we have not been able take into account (see \cref{sec:uninvestigated}), leaving room for future work.

%% file: section/appendix.tex
\appendix
\setcounter{equation}{0}
\renewcommand\theequation{A.\arabic{equation}}
\onecolumn
\renewcommand\thefigure{\thesection.\arabic{figure}}
\setcounter{figure}{0}

\section{Background on Gaussian Process Regression \& Its Variational Approximations}
In this section, we briefly provide additional background on GPR, SGPR, and SVGP.

\subsection{Gaussian Process Regression}\label{app:gpr}

Recalling our setup, we assume we have observed a dataset containing $N$ observations $(X, \vy) = \{x_n, y_n\}_{n=1}^N$, with $x_n \in \Xspace$, and $y_n \in \Reals$.
We assume $y_n = f(x_n) + \epsilon_n$, where the $\epsilon_n \sim \mathcal{N}(0, \sigma_n^2)$ are independent and identically distributed with noise variance $\sigma_n^2$.
We place a Gaussian process prior over $f$, writing $f|\vtheta \sim \GP(0, k_{\vtheta})$, where $k_{\vtheta}: \Xspace \times \Xspace \to \Reals$ is a covariance function with hyperparameters $\vtheta$.\footnote{Note that we have assumed a zero mean function. As we normalise our datasets, this does not present a significant issue for us.}
We write $K_{XX}$ to denote the $N\times N$ Gram matrix defined by this kernel applied to the data $X$, given by $[K_{XX}]_{ij} = k(x_i, x_j)$ (where we have omitted the dependence on $\vtheta$ for notational clarity).
This model therefore implies a $\mathcal{N}(\mathbf{0}, K_{XX})$ prior distribution over $\mathbf{f} = f(X)$, i.e., $\mathbf{f} \sim \mathcal{N}(0, K_{XX})$.
In combination with the likelihood $p(y|f(x)) = \mathcal{N}(0, \sigma^2)$ implied by our model, we can write $\vy | X, \vtheta \sim \mathcal{N}(\mathbf{0}, K_{XX} + \sigma_n^2 I_N)$, where $I_N$ is an identity matrix of size $N \times N$.
This leads to the log marginal likelihood (LML), which we use to optimise hyperparameters:
\begin{align}
    \log p(\vy|X,\,\vtheta) &= \log \mathcal{N}(\vy|\mathbf{0}, K_{XX} + \sigma_n^2 I_N) \\
    &= - \frac{N}{2} \log(2\pi) - \frac{1}{2}\vy^\top (K_{XX} + \sigma_n^2 I_N)^{-1}\vy - \frac{1}{2}\log |K_{XX} + \sigma_n^2 I_N|.
\end{align}
Note that the log marginal likelihood involves a trade-off between a ``data fit'' term (the quadratic term), and a log determinant ``complexity penalty,'' which makes it a suitable objective for optimising the kernel hyperparameters \citep[see][for a more in-depth discussion]{rasmussen2006gaussian}.\footnote{Again, we note that in some cases \citep{oberpromises21}, the LML is susceptible to overfitting; however, this does not apply here, where we have far fewer hyperparameters than datapoints.}

Furthermore, for a test point $x^*$, we can write 
\begin{align}
    \begin{pmatrix}
        \vy \\ 
        f(x^*)
    \end{pmatrix}
    \sim \mathcal{N}\left(
    \mathbf{0}, \,
    \begin{pmatrix}
        K_{XX} + \sigma_n^2 I_N & k_{Xx^*} \\
        k_{x^*X} & k(x^*,x^*)
    \end{pmatrix}
    \right),
\end{align}
where we use $k_{x^*X}$ to denote the row vector given by $[k(x^*, x_j)]_{j=1}^N$, and $k_{Xx^*}$ to denote its transpose column vector.
By using standard Gaussian conditioning rules, we can obtain predictions at $x^*$:
\begin{align}
    p(f(x^*) | \vy,\,X,\,\vtheta) = \mathcal{N}(f(x^*)|m(x^*), \hat{k}(x^*,x^*)),
\end{align}
where 
\begin{align}
    m(X^*) &= k_{x^*X}(K_{XX} + \sigma_n^2 I_N)^{-1}\vy, \\
    \hat{k}(x^*, x^*) &= k(x^*, x^*) - k_{x^*X}(K_{XX} + \sigma_n^2 I_N)^{-1}k_{Xx^*}.
\end{align}
We note that both computing the LML and computing predictions incur an $O(N^3)$ computational cost and $O(N^2)$ memory cost, due to the storage and computation of the inverse of $K_{XX} + \sigma_n^2 I_N$.

\subsection{Kernels}\label{app:kernels}

We briefly discuss the classes of kernels that appear in this work; we refer the reader to \citet{rasmussen2006gaussian} for a more detailed treatment of kernels.
The first, and most prevalent kernel in this work, is the squared exponential (SE) kernel.
In this work, we make use of the automatic relevance determination (ARD) version of the kernel, which allows the model to effectively remove irrelevant inputs.
Its form for $x, x' \in \Reals^D$ is given by
\begin{align}
    k(x, x') = \sigma_f^2 \exp\left(-\frac{1}{2}\sum_{d=1}^D \frac{(x_d - x'_d)^2}{l_d^2}\right),
\end{align}
where $\sigma_f^2$ is the signal variance, and $\{l_d\}_{d=1}^D$ are the lengthscales, leading to $\vtheta = \{\sigma_f, \{l_d\}_{d=1}^D, \sigma_n\}$.
For a GP with the squared exponential kernel, the posterior mean and samples from both the prior and posterior will be almost surely infinitely differentiable (i.e., smooth).

The next class of kernels we consider are the Mat\'{e}rn family of kernels.
This family is defined with a smoothness parameter $\nu > 0$, such that
\begin{align}
    k_\nu(x, x') = \sigma_f^2\frac{2^{1-\nu}}{\Gamma(\nu)}\left(\sqrt{2\nu}\|x - x'\|\right)^\nu K_{\nu}\left(\sqrt{2\nu}\|x-x'\|\right),
\end{align}
where $\Gamma(\cdot)$ is the gamma function, $K_{\nu}(\cdot)$ is the modified Bessel function of the second kind, and $\sigma_f^2$ is again the signal variance.
The Mat\'{e}rn family has the property that a GP using a kernel with parameter $\nu$ will have a posterior mean that is $\lfloor \nu \rfloor$ times differentiable.
Moreover, the kernel converges to the squared exponential kernel as $\nu \rightarrow \infty$, and it can be written in terms of elementary functions for half-integer values of $\nu$.
We extend the standard Mat\'{e}rn kernel to make use of ARD, leading again to a GP with hyperparameters $\vtheta = \{\sigma_f, \{l_d\}_{d=1}^D, \sigma_n\}$.

The final kernel we consider is the arc-cosine kernel from \citet{cho2009kernel}.
This kernel results from taking the infinite-width limit of a single-layer Bayesian neural network, resulting in 
\begin{align}
    k_n(x, x') = \frac{\sigma_f^2}{\pi} \|x\|^n \|x'\|^n J_n(\phi),
\end{align}
where $\phi = \arccos(x^\top x'/\|x\|\|x'\|)$, and $n$ is the order of the kernel.
Different orders correspond to different activation functions, with the following forms of $J_n(\cdot)$:
\begin{align}
    J_0(\phi) &= \pi - \phi, \\
    J_1(\phi) &= \sin(\phi) + (\pi - \phi) \cos(\phi), \\
    J_2(\phi) &= 3 \sin(\phi) \cos(\phi) + (\pi - \phi) (1 + 2\cos^2(\phi)).
\end{align}
For $n = 0$, the kernel corresponds to the Heaviside activation function; for $n=1$, it corresponds to the ReLU activation function; and for $n=2$ it corresponds to the half-quadratic activation function, i.e., $x \mapsto\max(0,\, \mathrm{sign}(x) x^2)$.
Note that, following \citet{gpflow2017}, we also include learnable input weight and bias variances $\{\sigma_{w,d}^2\}_{d=1}^D$ and $\sigma_b^2$, which means that we effectively map all pairs of inputs using
\begin{align}
    \langle x, x'\rangle = \sum_{d=1}^D \sigma_{w, d}^2 x_d x_d' + \sigma_b^2
\end{align}
before applying the standard arc-cosine kernel.
We note that this mapping causes no issues, as all the necessary computations for the kernel can achieved through inner products between points.
Therefore, for a GP with the arc-cosine kernel, we have the following trainable hyperparameters: $\{\sigma_f, \{\sigma_{w, d}\}_{d=1}^D, \sigma_b, \sigma_n\}$.

\subsection{Sparse Gaussian Process Regression}\label{app:sgpr}

We now turn to describing the sparse Gaussian process regression (SGPR) approximation of \citet{titsias2009variational}.
This approximation relies on $M \leq N$ inducing variables $\vu$, which are the values of the function $f$ at inducing locations $Z = \{z_m\}_{m=1}^M$, so that $\vu = f(Z)$.
By learning an approximate posterior $q(\vu)$ over the inducing variables, we can form an approximate GP through
\begin{align}\label{eq:sgpr-var-factorise}
    q(f) = q(f_{\neq \vu}, \vu) = q(f_{\neq \vu}|\vu) q(\vu),
\end{align}
which can be learned by maximising the evidence lower bound (ELBO) $\mathcal{L}$, a lower bound to the LML:
\begin{align}
    \log p(\vy|X) &\geq \log p(\vy|X) - \mathrm{KL}[q(f) || p(f | \vy] \\
    &= \mathbb{E}_{q(f)}[\log p(\vy|\mathbf{f})] - \mathrm{KL}[q(f) || p(f)] \eqqcolon \mathcal{L}, \label{eq:svgp-elbo}
\end{align}
where the inequality comes from the fact that the Kullback-Leibler (KL) divergence is non-negative, and where we have suppressed dependence on the hyperparameters $\vtheta$.\footnote{We present the derivation here informally; for the interested reader, a formal measure theoretic treatment of variational inference for GPs can be found in \citet{matthews2016sparse}.}
Substituting \cref{eq:sgpr-var-factorise}, it is possible to show that the optimal form of $q(f_{\neq \vu}|\vu)$ is given by $p(f_{\neq \vu}|\vu)$, i.e., the true GP conditional.
Moreover, for regression with Gaussian likelihoods, \citet{titsias2009variational} showed that the optimal $q(\vu)$ for the ELBO can be derived in closed form as a Gaussian.
Substituting this optimal form, it is possible to show that the ELBO reduces to
\begin{align}\label{eq:sgpr_elbo}
    \mathcal{L} = -\frac{N}{2} \log 2\pi -\frac{1}{2}\log |Q_{XX} + \sigma_n^2 I_N| - \frac{1}{2}\vy^\top (Q_{XX} + \sigma_n^2 I_N)^{-1}\vy - \frac{1}{2\sigma_n^2}\mathrm{tr}(K_{XX} - Q_{XX}), 
\end{align}
where we have defined $Q_{XX} = K_{XZ}K_{ZZ}^{-1}K_{ZX}$.
We note that this bound can be computed in $O(N^2M + M^3)$, and requires $O(NM + M^2)$ memory
Moreover, the ELBO is tight enough that it can be used to learn both the inducing locations \emph{and} the model hyperparameters.
Furthermore, \citet{titsias_variational_2014} showed that you can compute an upper bound on the LML in the same time and memory as follows:
\begin{align}\label{eq:sgpr-upperbound}
    \mathcal{L}_{\mathrm{upper}} = -\frac{N}{2}\log 2\pi - \frac{1}{2}\vy^\top (Q_{XX} + \sigma^2_n I_N + t I_N)^{-1}\vy - \frac{1}{2} \log |Q_{XX} + \sigma_n^2 I_N|,
\end{align}
where we have defined $t = \mathrm{tr}(K_{XX} - Q_{XX})$.
Finally, we can compute predictions by integrating out the inducing variables to obtain
\begin{align}\label{eq:sgpr-pred}
    q(f(x^*)) = \mathcal{N}(f(x^*)|m(x^*), \hat{k}(x^*,x^*)),
\end{align}
where
\begin{align}
    \hat{m}(x^*) &= k_{x^*Z}K_{ZZ}^{-1}K_{ZX}(Q_{XX} + \sigma_n^2 I_N)^{-1}\vy \\
    \hat{k}(x^*,x^*) &= k_{x^*x^*} - k_{x^*Z}K_{ZZ}^{-1}K_{ZX}(Q_{XX} + \sigma_n^2 I_N)^{-1}K_{ZX}K_{ZZ}^{-1}k_{Zx^*}.
\end{align}

\subsection{The Stochastic Variational Gaussian Process}\label{app:svgp}
Considering more recent advances for sparse variational inference, \citet{hensman2013gaussian} introduced a version of the ELBO that could be minibatched, resulting in a stochastic estimate of the ELBO, resulting in a stochastic variational Gaussian process (SVGP).
By retaining an explicit form of the approximate posterior, i.e., $q(\vu) = \mathcal{N}(\mathbf{m}, \mathbf{S})$, where $\mathbf{m}$ and $\mathbf{S}$ are trainable variational parameters, we can return to the ELBO of \cref{eq:svgp-elbo}:
\begin{align}
    \mathcal{L} = \mathbb{E}_{q(f)}[\log p(\vy|\mathbf{f})] - \mathrm{KL}[q(f)||p(f)].
\end{align}
Assuming a likelihood that factorises across datapoints, after some manipulation we can arrive at the following:
\begin{align}
    \mathcal{L} = \sum_{n=1}^N \mathbb{E}_{q(f(x_i))}[\log p(y_i | f(x_i))] - \mathrm{KL}[q(\vu) || p(\vu)].
\end{align}
The first term can be straightforwardly minibatched, and the expectation can be computed in closed form for Gaussian likelihoods.
Moreover, the second term can also be computed in closed form given a Gaussian approximate posterior.
Finally predictions are achieved through
\begin{align}
    q(f(x^*)) = \mathcal{N}(f(x^*)|m(x^*), \hat{k}(x^*,x^*)),
\end{align}
where
\begin{align}
    \hat{m}(x^*) &= k_{x^*Z}K_{ZZ}^{-1}\mathbf{m} \\
    \hat{k}(x^*,x^*) &= k_{x^*x^*} - k_{x^*Z}K_{ZZ}^{-1}(K_{ZZ} - \mathbf{S})K_{ZZ}^{-1}k_{Zx^*}.
\end{align}
We note that minibatching comes at the cost of no longer being able to use standard L-BFGS implementations (due to the stochasticity it introduces), as well as adding significantly more parameters to the optimisation problem, as $\mathbf{m}$ and $\mathbf{S}$ are no longer integrated out.

\section{SGPR Baseline Procedure}
In this section, we give a detailed account of our baseline procedure, including how we improved its numerical stability.
We begin by providing a pseudocode algorithm for our procedure in \cref{alg:sgpr-baseline}.

\begin{algorithm}
\caption{Training Procedure for SGPR\label{alg:sgpr-baseline}}
\begin{algorithmic}
    \STATE Given: Data $(X,\vy)$, initial parameters $\theta_0$.
    \STATE $\mathbf{M} \leftarrow \{10, 20, 50, 100, \dots, 10\_000\}$
    \FOR{$M$ in $\mathbf{M}$}
        \IF{$M > \mathrm{int}(0.8N)$}
            \BREAK
        \ENDIF
        \STATE initialise SGPR model with $M$ inducing points
        \STATE $\theta \leftarrow \theta_0$
        \STATE $Z \leftarrow \mathrm{greedy}\text{-}\mathrm{var}(\theta, M)$
        \FOR{epoch $\leftarrow$ 1 to 20}
            \STATE $\theta' \leftarrow$ maximise ELBO w.r.t. $\theta$
            \STATE $\mathrm{ELBO} \leftarrow \mathrm{ELBO}(\theta', Z)$
            \STATE $Z' \leftarrow \mathrm{greedy}\text{-}\mathrm{var}(\theta', M)$
            \STATE $\mathrm{ELBO}' \leftarrow \mathrm{ELBO}(\theta', Z')$
            \IF{$\mathrm{ELBO}' < \mathrm{ELBO}$} 
                \BREAK
            \ENDIF
            \STATE $\theta,\,Z \leftarrow \theta',\,Z'$
        \ENDFOR
        \STATE Record ELBO, upper bound, and predictive metrics (RMSE, NLPD)
    \ENDFOR
\end{algorithmic}
\end{algorithm}
We first note that our procedure requires identical inputs as exact GP regression, making it fully automatic.
In the algorithm, we begin by constructing a list of numbers of inducing points, increasing approximately logarithmically from 10 to 10,000.
The outermost loop involves iterating through this list, thereby gradually increasing the amount of computation needed to obtain a Pareto front describing the accuracy-compute trade-off.
Within the outermost loop, we first check whether the selected $M$ is greater than 0.8 times the number of datapoints, and stop the procedure if so: this is the point at which we deem there will likely be no real advantage to using SGPR over exact GPR.
Following that, we initialise the SGPR model with the given initial parameters $\theta_0$ and the inducing point locations $Z$ given by the greedy variance procedure from \citet{burt2020convergence}.
Following this initialisation, we alternate maximising the ELBO with respect to $\theta$ with re-initialising the inducing locations, up to 20 times, again following the procedure from \citet{burt2020convergence}.
At each iteration of this inner loop, we check whether the greedy variance selection has led to an improved ELBO; if not, we terminate the procedure, and move to the next $M$.
At the end of our procedure for each $M$, we record the relevant metrics, including ELBOs, upper bounds, and the predictive RMSEs and NLPDs.

\subsection{Improving Numerical Stability}\label{app:numerical-stability}

After a close investigation, we found that making two changes to our above training routine ensured that our procedure was robust.
The first term concerned the term $\mathrm{tr}(K_{XX} - Q_{XX})$, which corresponds to the trace of the covariance of $p(\mathbf{f}|\vu)$.
As this term is the sum of variances, it should be non-negative.
However, in practice, an accumulation of numerical errors may cause it to be negative.
To address this, we correct the trace term to be zero in the ELBO computation if the errors in the trace are deemed to be relatively small.
If not, we we return a NaN.
We provide the precise procedure in \cref{alg:trace}.

\begin{algorithm}
\caption{Trace computation with numerical error fixing\label{alg:trace}}
\begin{algorithmic}

    \IF{$\mathrm{tr}(K_{XX} - Q_{XX}) < 0$}
        \STATE $\mathrm{tr}(K_{XX} - Q_{XX}) \leftarrow 0$
    \ELSE
        \RETURN $\mathrm{NaN}$
    \ENDIF
\end{algorithmic}
\end{algorithm}

As mentioned in the main text, for our second change, we modify the standard L-BFGS implementation to include restarts, where we clear the optimiser history and reset the Hessian approximation.
Such restarts are triggered when a function evaluation produces an error or returns a $\mathrm{NaN}$, which typically happens when the line search queries unrealistic hyperparameter values.
In this case, we hope that resetting the optimiser's internal state will cause it to search along a new, more promising direction in hyperparameter space.
This approach of resetting the L-BFGS optimiser is inspired by similar restart procedures in the literature \citep{rasmussen2010gaussian, zhu1997algorithm}.

We found that the combination of these modifications to the computation of the ELBO and its optimisation led to a robust implementation of our procedure where we no longer needed to tune upper bounds on hyperparameter values, meaning that we could run the same exact procedure on all datasets without fear of numerical failure.

\section{Recommendations for Stochastic Variational Gaussian Processes}\label{app:svgp-exp}
\begin{figure*}[ht]
     \centering
    \includegraphics[width=\textwidth]{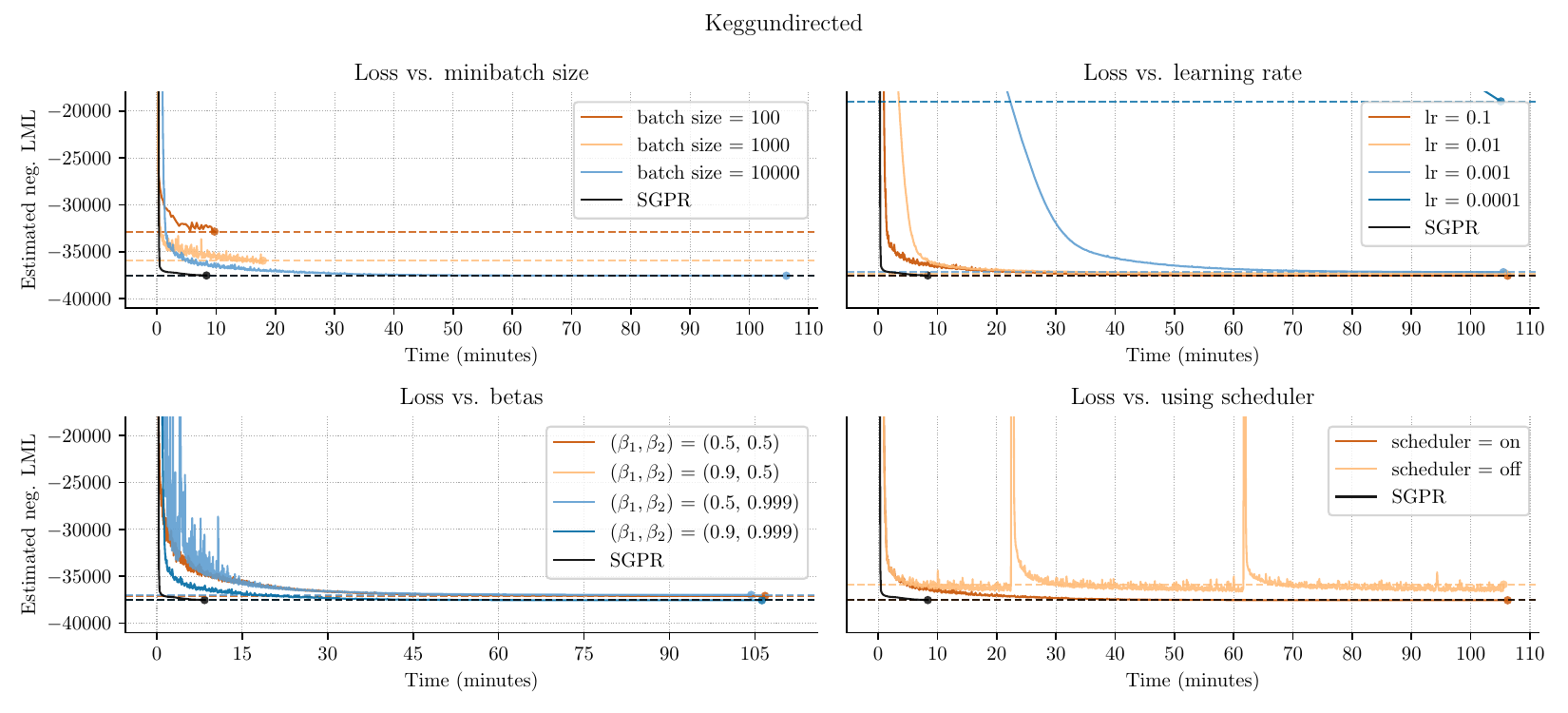}
    \caption{We plot training curves for SVGP with 1000 inducing points on \texttt{keggundirected} with various hyperparameter settings, changing from the optimal hyperparameter setting found using the described grid search. We also plot the SGPR mean from our proposed procedure (extracting the values for $M=1000$). \textbf{Top left}: Dependence on minibatch size. \textbf{Top right}: Dependence on learning rate. \textbf{Bottom left}: Dependence on optimiser momentum parameters $(\beta_1, \beta_2)$. \textbf{Bottom right}: Dependence on use of a scheduler.}
    \label{fig:svgp-losses}
\end{figure*}

In this section, we perform a hyperparameter search for SVGP trained on the \texttt{keggundirected} dataset with $M=1000$ inducing points.
We use the Adam \citep{kingma2014adam} optimiser, training for 20,000 gradient steps, and perform a grid search over the following hyperparameter settings:
\begin{itemize}
    \item batch size $\in \{100, 1000, 10,000\}$;
    \item learning rate $\in \{0.1, 0.01, 0.001, 0.0001\}$;
    \item $\beta_1 \in (0.5, 0.9)$;
    \item $\beta_2 \in (0.5, 0.999)$;
    \item using a learning rate scheduler versus no learning rate scheduler.
\end{itemize}
For the learning rate scheduler, we use a reduce-on-plateau scheduler with patience 10 (in epochs), factor 0.95, threshold 0, and minimum learning rate of 1e-6.
We found that the best-performing combination, in terms of final ELBO value, was batch size 10,000, learning rate 0.1, $(\beta_1, \beta_2) = (0.9, 0.999)$, using a learning rate scheduler.
We take this combination and ablate the hyperparameters one-by-one (paired for the $\beta$s), plotting the negative ELBO training curves in Fig.~\ref{fig:svgp-losses}.
We also compare to the mean ELBO training curves for SGPR with $M=1000$, taken from our main experiment (\cref{fig:near-exact-UCI}).

We first observe that no setting matches the speed of SGPR, and our best combination only just improves on its ELBO.
This is despite the fact that we allow the inducing locations to be trained for SVGP through gradient-based optimisation, whereas we fix the locations to the datapoints chosen by the greedy variance initialistion for SGPR.
In terms of hyperparameters, we see a clear dependence on minibatch size: increasing the minibatch size reduces noise in the update, which helps the optimisation \citep{wang2023joint}.
For learning rate, we see that increasing the learning rate helps the rate of convergence, but at the cost of more noisy updates.
For the $\beta$s, we see that the use of $\beta_1 = \beta_2 = 0.5$ dramatically improves the initial rate of convergence; however, the loss for the default values quickly catch up.
Nevertheless, we found that for lower values of learning rate (e.g., 0.01), $\beta_1 = \beta_2 = 0.5$ provides a better final ELBO, but this is negated by lowering the batch size.
This suggests that a non-default value of the $\beta$s might be beneficial in lower-noise situations for GPs; we leave this for future work.
Finally, we observe that using a learning rate scheduler is crucial when starting with this high of a learning rate.

Overall, despite the large grid search over hyperparameter values, we were not able to find hyperparameter settings that truly outcompete SGPR. 
Moreover, we observe that SVGPs should be trained for many iterations for competitive performance, and at high minibatch sizes.
While we make some recommendations for SVGP training, it is likely that different datasets will require different settings, and thus it may be difficult to find a universal recommended training procedure that performs as well as SGPR.

\section{Uninvestigated Issues}
\label{sec:uninvestigated}

\begin{figure*}[ht]
     \centering
    \includegraphics[width=\textwidth]{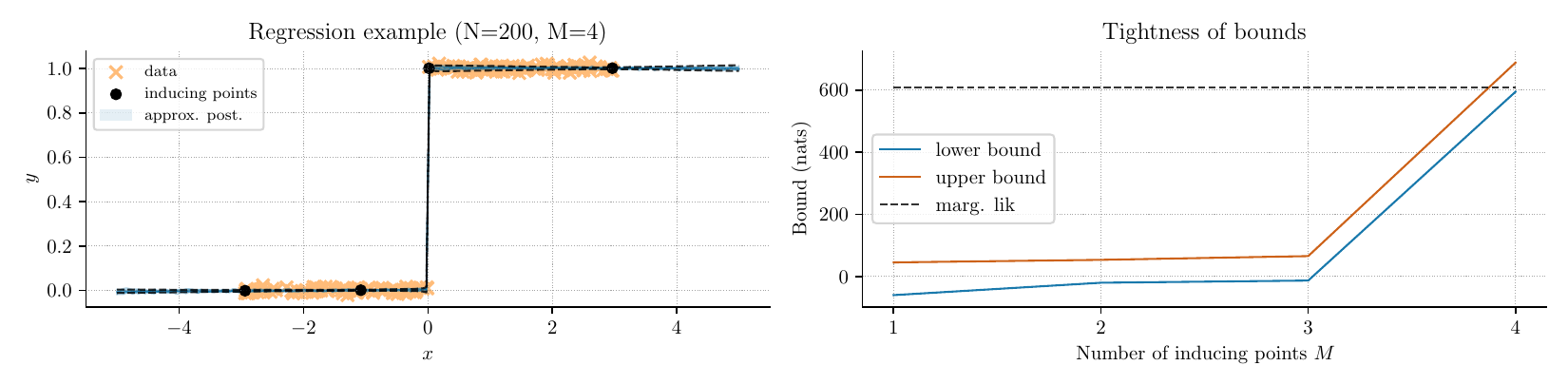}
    \caption{SGPR with an arc-cosine kernel (\cref{app:kernels}) on a step dataset. \textbf{Left}: Example approximate solution. \textbf{Right}: Upper and lower bounds on marginal likelihood with varying $M$. Note that different hyperparameters are found as $M$ increases. In contrast to the Mat\'{e}rn-$\frac{1}{2}$ kernel, the upper and lower bounds quickly converge.}
    \label{fig:sgpr-working-relu}
\end{figure*}

One uninvestigated effect in our main experiments is the interplay between kernel choice and approximation. 
For SGPR, the number of required inducing points depends strongly on the kernel choice. 
In our UCI experiments, we follow the literature and only investigate a fixed kernel (here, the squared exponential). 
However, in some cases, finding a more well-specified kernel can also make a model much cheaper to approximate. 
As briefly mentioned in the main text, an example of this is shown in \cref{fig:sgpr-working-relu}, where we use an arc-cosine kernel for step function data, which can then be approximated perfectly with only four inducing points, rather than the failure case in \cref{fig:sgpr-failing-toy}. 
Perhaps the current approach of investigating GP approximations in isolation of kernel search has fundamental barriers.

In this paper, we focused on the time cost of GP approximations, which is typically the most relevant computational concern in the literature.
However, it may also be the case that a practitioner may want to minimise FLOPs (for instance, for energy efficiency) or memory usage.
Indeed, as briefly discussed in the main text, the latter of these is often more relevant than time for a practitioner, as out-of-memory errors will immediately halt computation.
If these are concerns, it is possible to plot similar metrics against both increasing FLOPs and/or memory usage, to come to a determination of the Pareto frontier for a method.

Finally, we did not take into account the prediction time. 
During our training procedure we logged test set metrics, which took significant time, particularly for the recommended iterative GP procedure \citep{maddox2021reliablyaccurate}. 
We did not include this time in the measured training time curves, to simulate a situation where a practitioner continues to run until a specified point, at which point performance is measured. 
Perhaps test-time performance experiments should be included following similar recommendations as the ones we already make.

\section{Additional Experimental Details}

We use the SE ARD kernel for all of our timed performance evaluations (\cref{sec:experiments}).
For our SGPR experiments, we initialise the noise variance to 0.01, and initialise the lengthscales and signal variances to 1.
We place lower limits on the feasible values of hyperparameters of $10^{-5}$, and use an initial value of $10^{-10}$ for our adaptive jitter, allowing up to 10 Cholesky attempts (increasing by a factor of 10 for each successive attempt).
As the evaluation metrics have large discontinuities when we initialise a new model with increased $M$ (we reinitialise the hyperparameters from scratch to avoid getting stuck in local optima preferred by models with smaller $M$), we smooth the values by setting them equal to the last value before the discontinuity, until the ELBO catches up again (up to a tolerance).
Note that we can do this safely smooth this way, as we are guaranteed that increasing $M$ will monotonically improve the ELBO.

For the CG-based iterative GP, we use the GPyTorch \citep{gardner2018gpytorch} implementation of \citet{wang2019exact}.
We changed the default settings to match the recommendations from \citet{maddox2021reliablyaccurate}.
We initialise the noise variance to 0.1, and the lengthscales and kernel variance to 1.
We evaluate the test metrics every 5 iterations, but as mentioned do not count the time for evaluating test metrics in our plots for either our baseline or the iterative GP.

For all methods, we implement an 8-hour time cutoff for a single run.
We use 85/15\% train/test splits for each UCI dataset.
We additionally only time the training portions of the runs, thereby excluding the time it takes to evaluate predictive metrics.

For our SVGP experiments, we retain the same setup as for SGPR.
We initialise the noise variance to 0.1, and the kernel hyperparameters to 1.
In addition, we initalise $\textbf{m} = \textbf{0}$ and $\mathrm{chol}(\mathbf{S}) = I_M$.
We initialise the inducing locations using the same greedy variance initialisation as for SGPR, which is taken into account in the timings.
However, unlike for SGPR, we learn our inducing locations through optimisation.

The hardware for the experiments are machines with 4 NVIDIA A6000 GPUs and an AMD EPYC 7402P CPU. 
Each experiment was run on a single GPU with exclusive access to minimise interference.

\section{Figures}\label{app:fig}

\clearpage
\section{Individual Datasets}
\label{sec:uci-plots}

\begin{figure}[h]
  \centering
  \includegraphics[width=\textwidth]{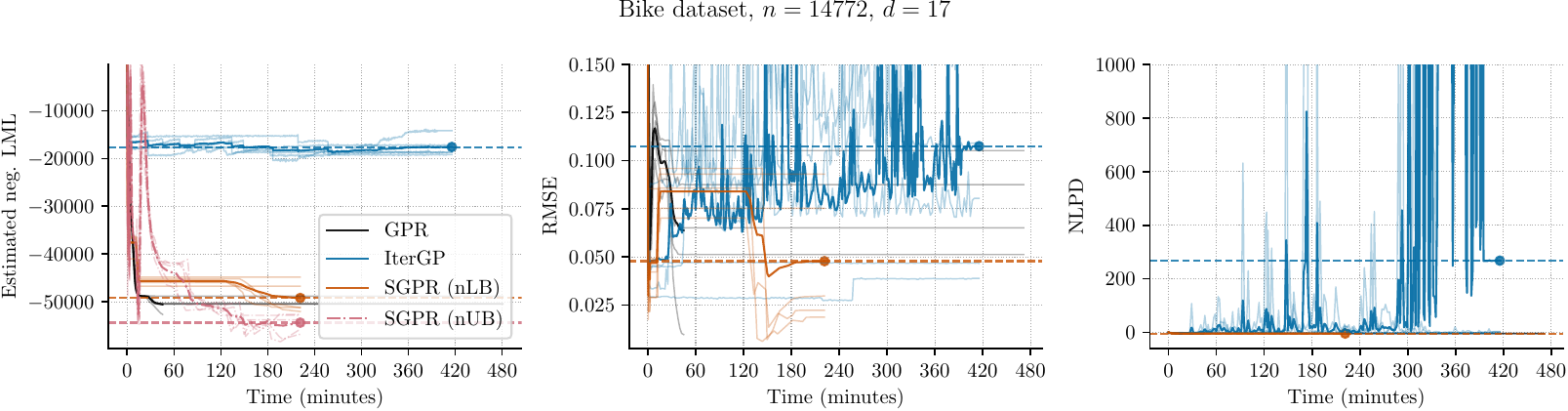}
  \caption{}
\end{figure}

\begin{figure}[h]
  \centering
  \includegraphics[width=\textwidth]{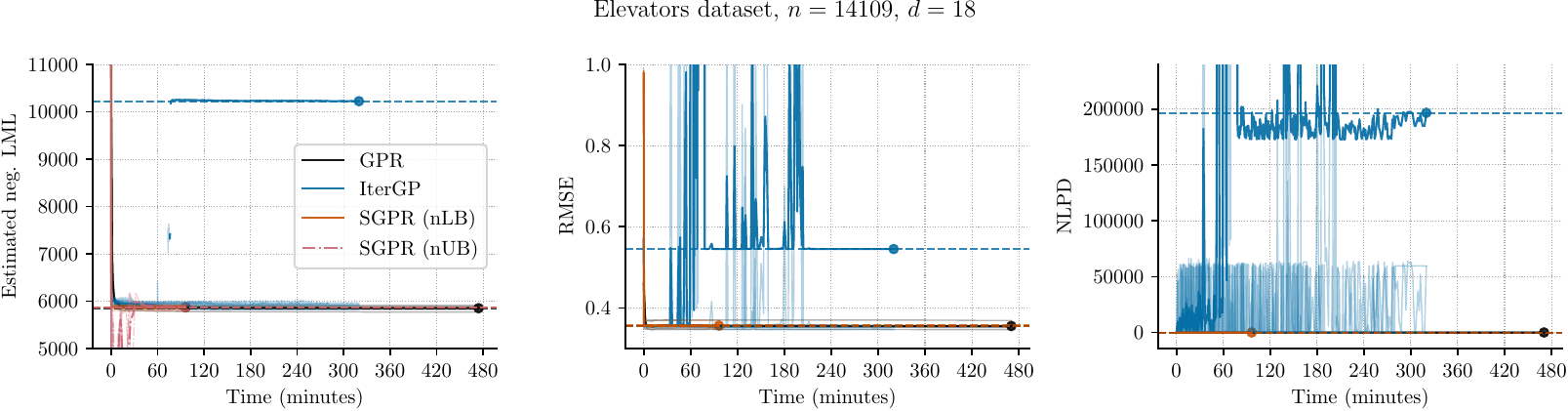}
  \caption{}
\end{figure}

\begin{figure}[h]
  \centering
  \includegraphics[width=\textwidth]{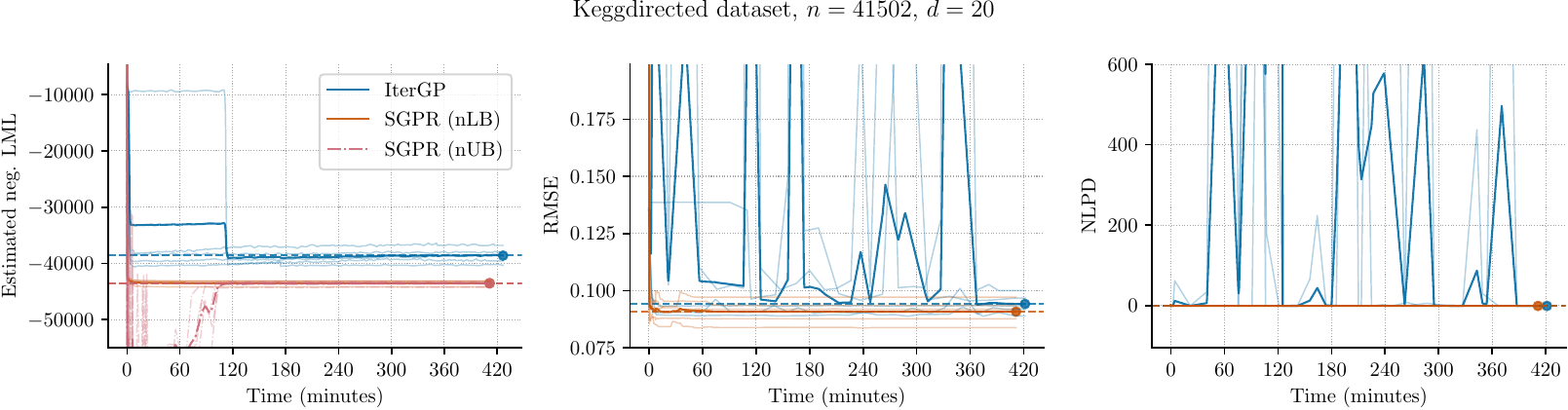}
  \caption{}
\end{figure}

\begin{figure}[h]
  \centering
  \includegraphics[width=\textwidth]{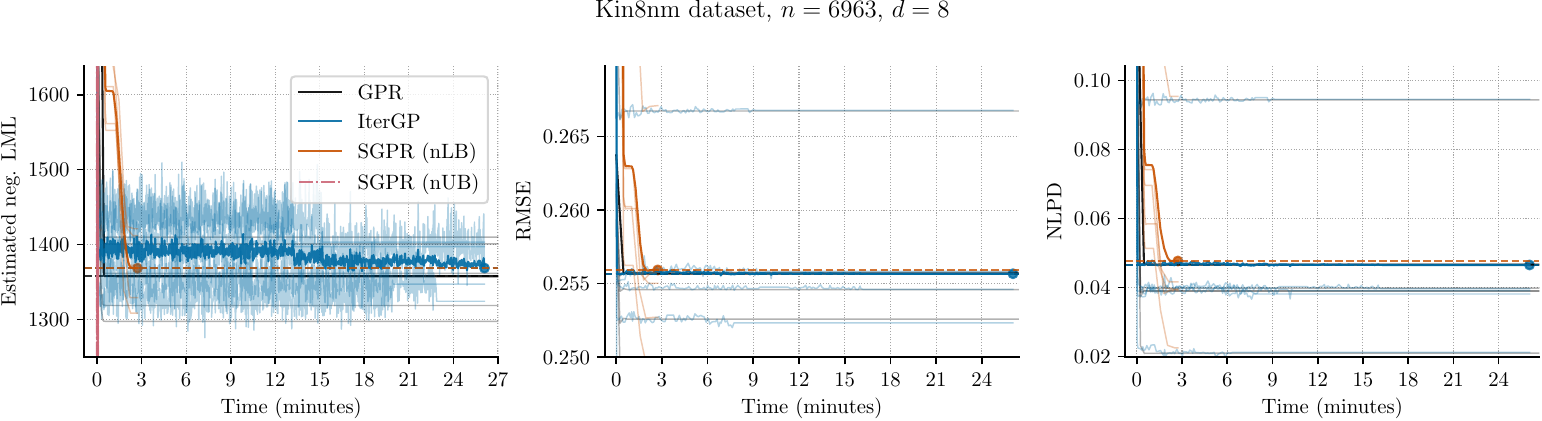}
  \caption{}
\end{figure}

\begin{figure}[h]
  \centering
  \includegraphics[width=\textwidth]{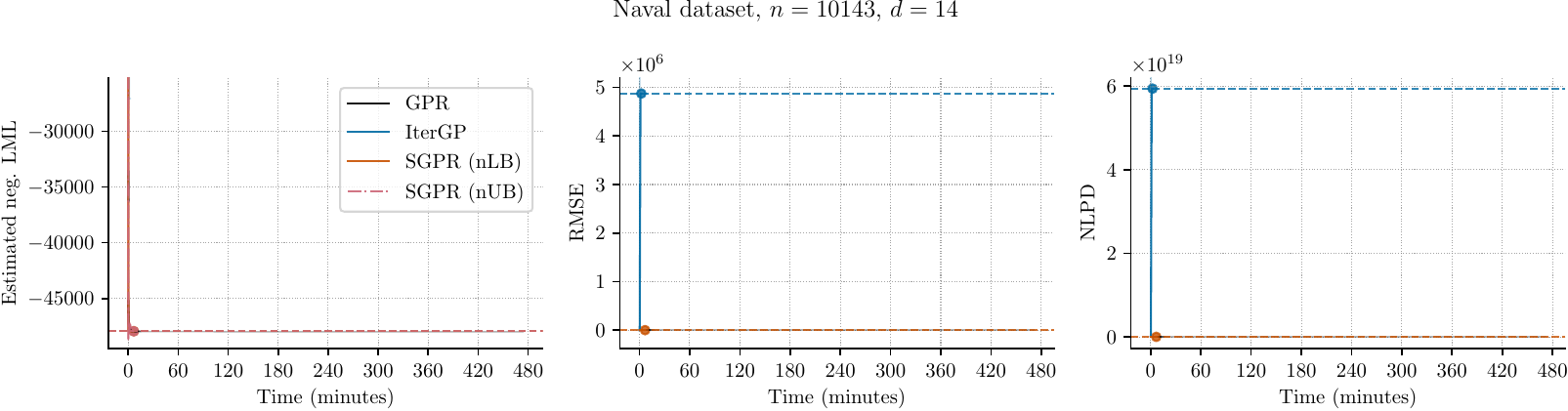}
  \caption{}
\end{figure}

\begin{figure}[h]
  \centering
  \includegraphics[width=\textwidth]{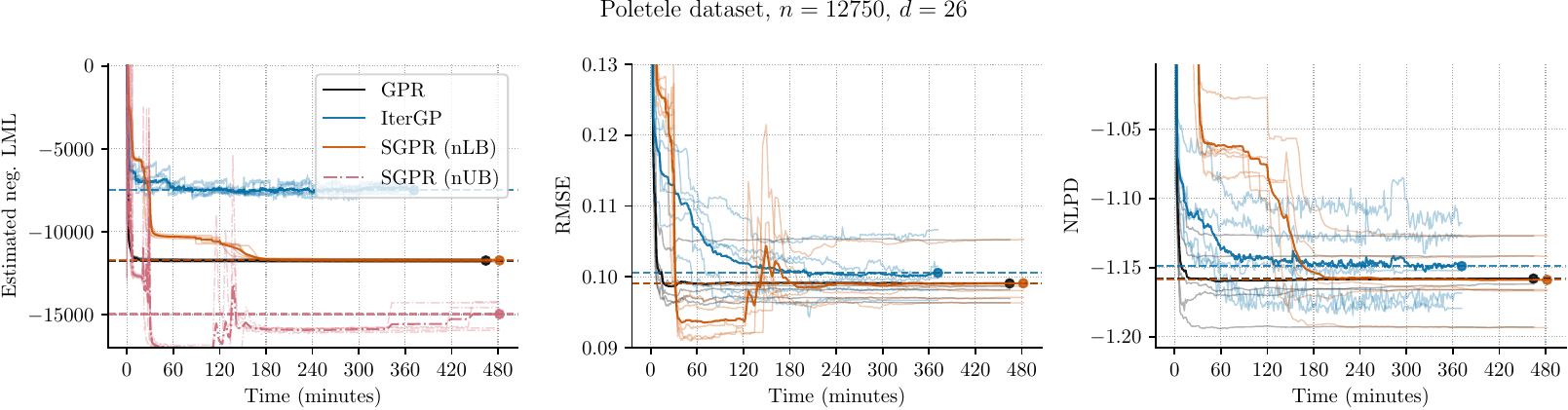}
  \caption{}
\end{figure}

\begin{figure}[h]
  \centering
  \includegraphics[width=\textwidth]{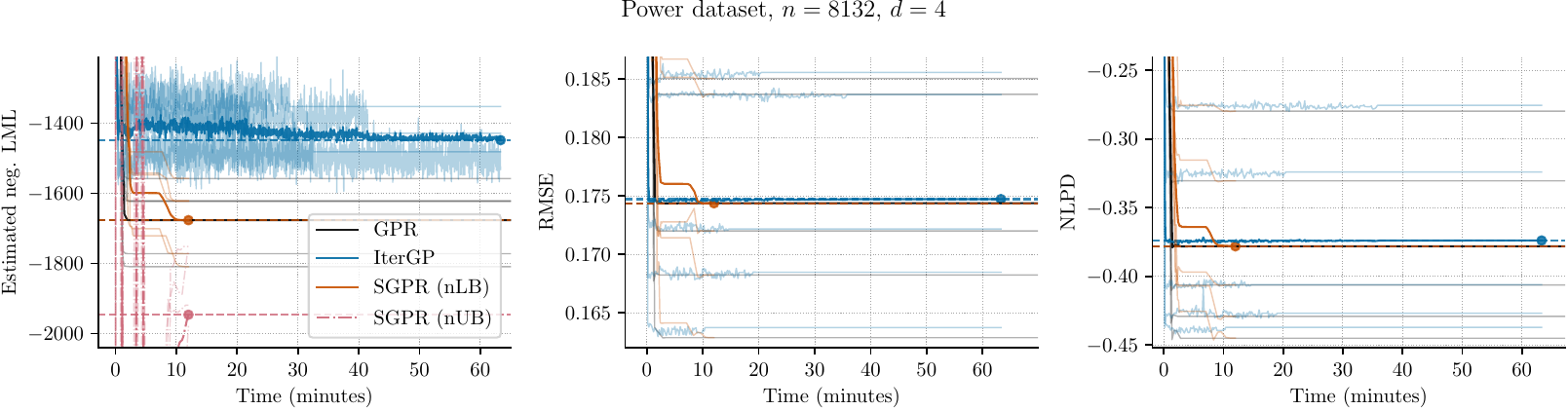}
  \caption{}
\end{figure}

\begin{figure}[h]
  \centering
  \includegraphics[width=\textwidth]{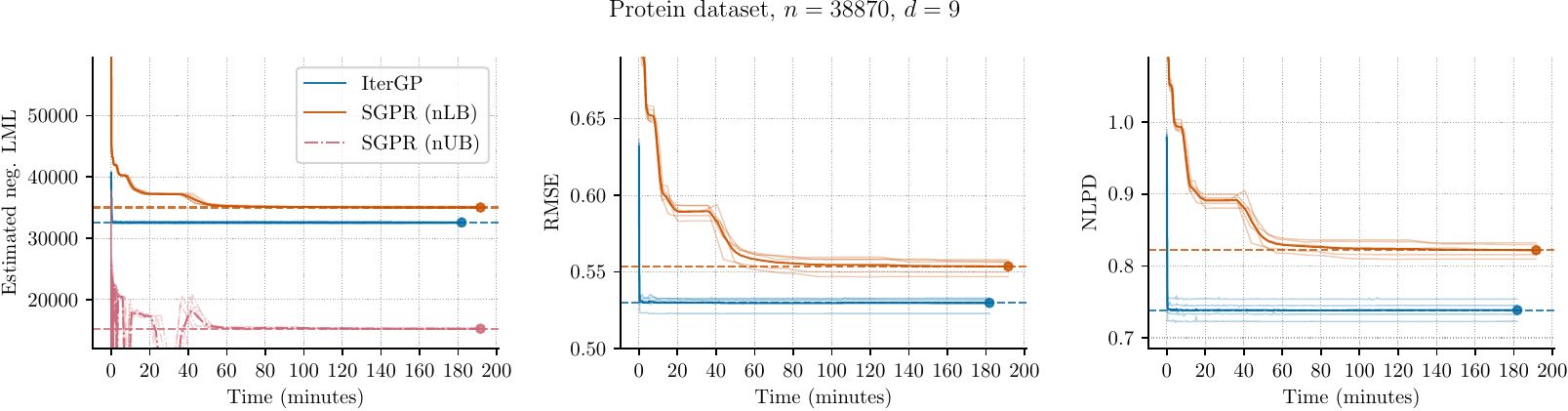}
  \caption{}
\end{figure}

\begin{figure}[h]
  \centering
  \includegraphics[width=\textwidth]{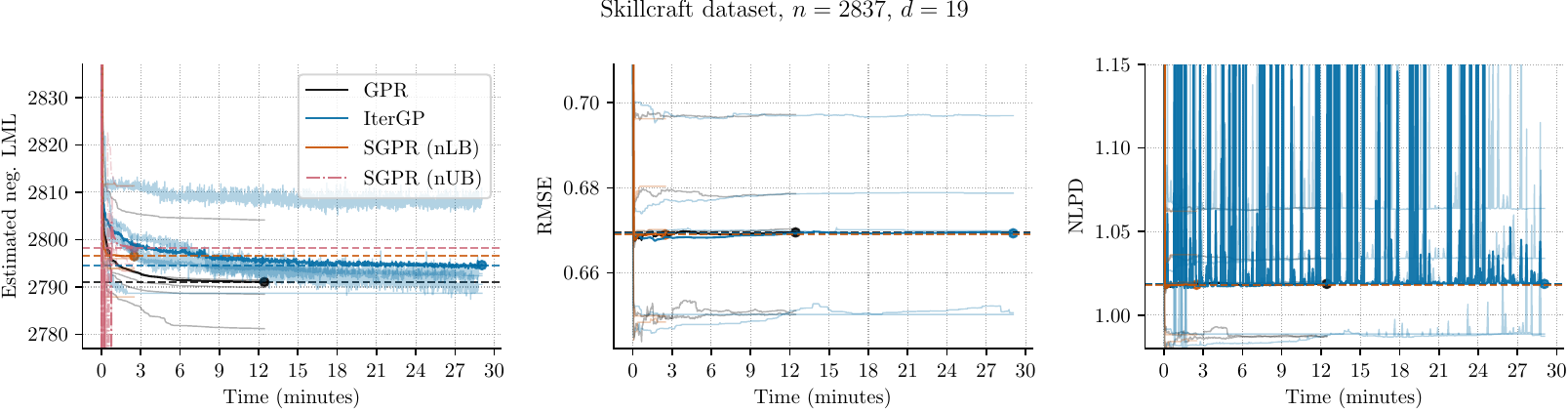}
  \caption{}
\end{figure}

\begin{figure}[h]
  \centering
  \includegraphics[width=\textwidth]{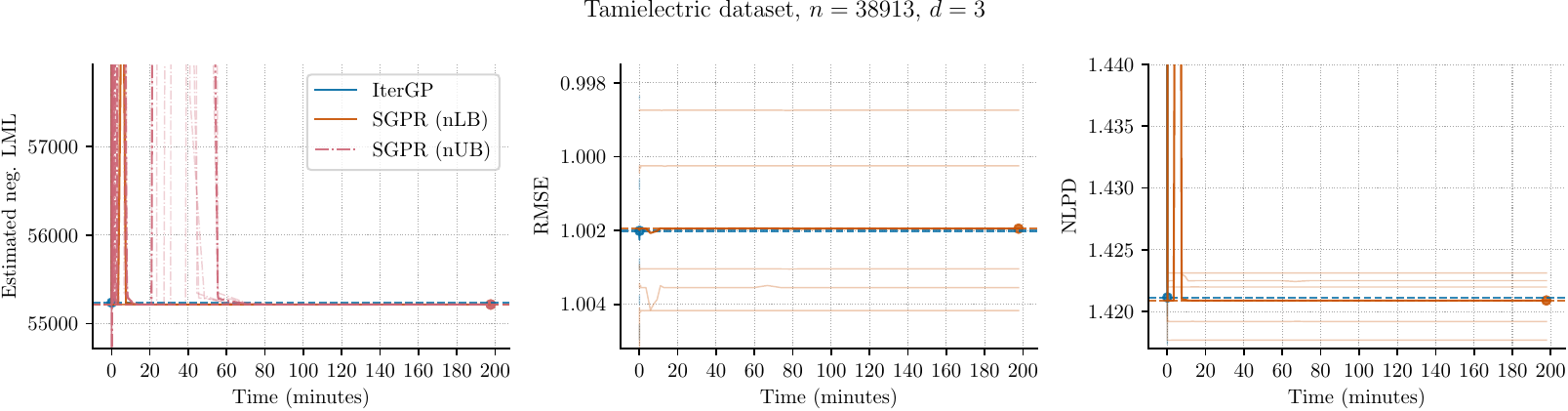}
  \caption{}
\end{figure}

\clearpage
\section{Tables}\label{app:tables}
In this section, we present tables of the metrics against runtime, averaged over five seeds.
Appropriate time checkpoints were selected using the SGPR procedure, as our procedure gives a natural point to check metrics: each time indicates the median time our procedure ran for before increasing the number of inducing points to the subsequent $M$ (however, the final column lists final values for all methods, irrespective of time and the final value of $M$, the latter of which depends on dataset size).
Therefore, we list the corresponding $M$ in the top row of the table.
For SGPR, tabulated metrics themselves correspond to the mean (over runs) of the final metrics for each $M$, and to the mean of the metrics for IterGP and GPR at the time checkpoints selected by SGPR.
For each dataset, we also provide the training set size, $n$, and the input dimension.
Finally, we also list metrics for linear regression and a constant mean prediction (i.e., taking the mean of the outputs as the prediction for all inputs) as sanity checks.
\begin{table}[h!]
\caption{LML approximations vs times} \label{tab:app:lmls}
\tiny
\begin{center}
\input{tables/losses}
\end{center}
\end{table}

\begin{table}[h]
\caption{RMSEs vs times} \label{tab:app:rmse}
\tiny
\begin{center}
\input{tables/rmse}
\end{center}
\end{table}

\begin{table}[h]
\caption{NLPDs vs times} \label{tab:app:nlpd}
\tiny
\begin{center}
\input{tables/nlpd}
\end{center}
\end{table}

%% file: tables/losses.tex
\begin{tabular}{clrrrrrrrrrr}
\hline
                                        Dataset                                         & Method     &    $m=\num{10}$ &      $\num{20}$ &      $\num{50}$ &      $\num{100}$ &      $\num{200}$ &      $\num{500}$ &      $\num{1000}$ &       $\num{2000}$ &       $\num{5000}$ &   final $M$ \\
\hline
      \hline \multirow{6}{*}{\shortstack[c]{\texttt{bike} \\ $n=14772$ \\ $d=17$}}      &            & $1.3\textup{s}$ & $2.4\textup{s}$ & $5.3\textup{s}$ &  $7.2\textup{s}$ & $11.2\textup{s}$ & $25.0\textup{s}$ &  $56.6\textup{s}$ &  $180.0\textup{s}$ &  $955.9\textup{s}$ &       final \\
                                                                                        & IterGP     &         $-5162$ &         $-9882$ &        $-11930$ &         $-14498$ &         $-15696$ &         $-15735$ &          $-15699$ &           $-16625$ &           $-16453$ &    $-17591$ \\
                                                                                        & SGPR       &         $-2081$ &         $-6286$ &        $-19817$ &         $-20999$ &         $-22134$ &         $-28198$ &          $-33367$ &           $-37599$ &           $-45641$ &    $-49212$ \\
                                                                                        & GPR        &          $4466$ &          $4213$ &          $3576$ &           $3142$ &           $2254$ &           $-852$ &           $-5136$ &           $-24254$ &           $-48744$ &    $-50551$ \\
                                                                                        & Linear     &              -- &              -- &              -- &               -- &               -- &               -- &                -- &                 -- &                 -- &     $11243$ \\
                                                                                        & Mean Pred. &              -- &              -- &              -- &               -- &               -- &               -- &                -- &                 -- &                 -- &     $20961$ \\
   \hline \multirow{6}{*}{\shortstack[c]{\texttt{elevators} \\ $n=14109$ \\ $d=18$}}    &            & $1.0\textup{s}$ & $3.4\textup{s}$ & $5.1\textup{s}$ &  $7.1\textup{s}$ & $13.0\textup{s}$ & $28.7\textup{s}$ &  $62.7\textup{s}$ &  $166.8\textup{s}$ &  $770.8\textup{s}$ &       final \\
                                                                                        & IterGP     &          $8316$ &          $6521$ &          $6217$ &           $6114$ &           $6045$ &           $6021$ &            $6016$ &             $6003$ &             $5954$ &     $10224$ \\
                                                                                        & SGPR       &          $9057$ &          $7571$ &          $7263$ &           $6585$ &           $6147$ &           $5904$ &            $5886$ &             $5875$ &             $5872$ &      $5871$ \\
                                                                                        & GPR        &         $12037$ &         $11778$ &         $11582$ &          $11371$ &          $10716$ &           $9474$ &            $7939$ &             $6323$ &             $5890$ &      $5853$ \\
                                                                                        & Linear     &              -- &              -- &              -- &               -- &               -- &               -- &                -- &                 -- &                 -- &      $9928$ \\
                                                                                        & Mean Pred. &              -- &              -- &              -- &               -- &               -- &               -- &                -- &                 -- &                 -- &     $20020$ \\
  \hline \multirow{6}{*}{\shortstack[c]{\texttt{keggdirected} \\ $n=41502$ \\ $d=20$}}  &            & $1.6\textup{s}$ & $3.3\textup{s}$ & $5.0\textup{s}$ &  $8.1\textup{s}$ & $15.5\textup{s}$ & $63.6\textup{s}$ & $278.2\textup{s}$ & $1023.8\textup{s}$ & $4854.8\textup{s}$ &       final \\
                                                                                        & IterGP     &         $-8472$ &         $-8496$ &         $-8518$ &          $-8560$ &          $-8661$ &          $-9369$ &          $-33124$ &           $-33231$ &           $-33019$ &    $-38618$ \\
                                                                                        & SGPR       &        $-20247$ &        $-30272$ &        $-36451$ &         $-39204$ &         $-41243$ &         $-42714$ &          $-43218$ &           $-43444$ &           $-43510$ &    $-43511$ \\
                                                                                        & Linear     &              -- &              -- &              -- &               -- &               -- &               -- &                -- &                 -- &                 -- &      $4834$ \\
                                                                                        & Mean Pred. &              -- &              -- &              -- &               -- &               -- &               -- &                -- &                 -- &                 -- &     $58889$ \\
 \hline \multirow{6}{*}{\shortstack[c]{\texttt{keggundirected} \\ $n=54066$ \\ $d=27$}} &            & $2.0\textup{s}$ & $4.3\textup{s}$ & $7.4\textup{s}$ & $12.6\textup{s}$ & $21.1\textup{s}$ & $88.5\textup{s}$ & $485.6\textup{s}$ & $1498.5\textup{s}$ & $8298.0\textup{s}$ &       final \\
                                                                                        & IterGP     &          $1160$ &          $1012$ &           $812$ &            $480$ &            $-65$ &          $-2522$ &          $-35211$ &           $-35445$ &           $-36061$ &    $-36081$ \\
                                                                                        & SGPR       &        $-25590$ &        $-32020$ &        $-34175$ &         $-35926$ &         $-36454$ &         $-37084$ &          $-37477$ &           $-37663$ &           $-37695$ &    $-37694$ \\
                                                                                        & Linear     &              -- &              -- &              -- &               -- &               -- &               -- &                -- &                 -- &                 -- &    $-20227$ \\
                                                                                        & Mean Pred. &              -- &              -- &              -- &               -- &               -- &               -- &                -- &                 -- &                 -- &     $76716$ \\
     \hline \multirow{6}{*}{\shortstack[c]{\texttt{kin40k} \\ $n=34000$ \\ $d=8$}}      &            & $0.6\textup{s}$ & $1.3\textup{s}$ & $2.7\textup{s}$ &  $4.5\textup{s}$ &  $8.7\textup{s}$ & $17.7\textup{s}$ &  $44.5\textup{s}$ &  $149.3\textup{s}$ &  $680.0\textup{s}$ &       final \\
                                                                                        & IterGP     &         $-6016$ &         $-6715$ &         $-8000$ &          $-9713$ &         $-13647$ &         $-17871$ &          $-19823$ &           $-19802$ &           $-19765$ &    $-19446$ \\
                                                                                        & SGPR       &         $45202$ &         $41835$ &         $35306$ &          $30004$ &          $22153$ &          $11361$ &            $2587$ &            $-7167$ &           $-19378$ &    $-26659$ \\
                                                                                        & Linear     &              -- &              -- &              -- &               -- &               -- &               -- &                -- &                 -- &                 -- &     $48237$ \\
                                                                                        & Mean Pred. &              -- &              -- &              -- &               -- &               -- &               -- &                -- &                 -- &                 -- &     $48244$ \\
      \hline \multirow{6}{*}{\shortstack[c]{\texttt{kin8nm} \\ $n=6963$ \\ $d=8$}}      &            & $0.6\textup{s}$ & $1.7\textup{s}$ & $3.4\textup{s}$ &  $4.2\textup{s}$ &  $5.8\textup{s}$ &  $8.9\textup{s}$ &  $16.1\textup{s}$ &   $38.3\textup{s}$ &                 -- &       final \\
                                                                                        & IterGP     &          $1831$ &          $1410$ &          $1388$ &           $1391$ &           $1388$ &           $1385$ &            $1396$ &             $1393$ &                 -- &      $1369$ \\
                                                                                        & SGPR       &          $7695$ &          $7074$ &          $5489$ &           $5104$ &           $3777$ &           $2835$ &            $2109$ &             $1605$ &                 -- &      $1369$ \\
                                                                                        & GPR        &          $2315$ &          $2139$ &          $1875$ &           $1800$ &           $1636$ &           $1524$ &            $1385$ &             $1358$ &                 -- &      $1358$ \\
                                                                                        & Linear     &              -- &              -- &              -- &               -- &               -- &               -- &                -- &                 -- &                 -- &      $8018$ \\
                                                                                        & Mean Pred. &              -- &              -- &              -- &               -- &               -- &               -- &                -- &                 -- &                 -- &      $9880$ \\
     \hline \multirow{6}{*}{\shortstack[c]{\texttt{naval} \\ $n=10143$ \\ $d=14$}}      &            & $1.9\textup{s}$ & $2.5\textup{s}$ & $4.3\textup{s}$ &  $6.3\textup{s}$ &  $9.0\textup{s}$ & $13.1\textup{s}$ &  $28.6\textup{s}$ &   $78.6\textup{s}$ &                 -- &       final \\
                                                                                        & IterGP     &        $-35406$ &        $-36051$ &        $-36182$ &            $nan$ &            $nan$ &            $nan$ &             $nan$ &              $nan$ &                 -- &    $-36172$ \\
                                                                                        & SGPR       &         $-7366$ &        $-20908$ &        $-44523$ &         $-47782$ &         $-47832$ &         $-47887$ &          $-47893$ &           $-47946$ &                 -- &    $-47946$ \\
                                                                                        & GPR        &        $-45527$ &        $-45642$ &        $-45945$ &         $-46282$ &         $-46747$ &         $-46906$ &          $-47271$ &           $-47566$ &                 -- &    $-47957$ \\
                                                                                        & Linear     &              -- &              -- &              -- &               -- &               -- &               -- &                -- &                 -- &                 -- &      $4987$ \\
                                                                                        & Mean Pred. &              -- &              -- &              -- &               -- &               -- &               -- &                -- &                 -- &                 -- &     $14392$ \\
      \hline \multirow{6}{*}{\shortstack[c]{\texttt{pol} \\ $n=12750$ \\ $d=26$}}       &            & $1.0\textup{s}$ & $2.5\textup{s}$ & $3.8\textup{s}$ &  $8.0\textup{s}$ & $12.2\textup{s}$ & $92.4\textup{s}$ & $310.3\textup{s}$ & $1660.5\textup{s}$ & $7976.3\textup{s}$ &       final \\
                                                                                        & IterGP     &          $1020$ &         $-1425$ &         $-2183$ &          $-2809$ &          $-2880$ &          $-4025$ &           $-6324$ &            $-7006$ &            $-7577$ &     $-7496$ \\
                                                                                        & SGPR       &         $11798$ &          $7125$ &          $7113$ &           $5861$ &           $3703$ &           $-488$ &           $-3669$ &            $-7498$ &           $-10866$ &    $-11733$ \\
                                                                                        & GPR        &           $672$ &           $447$ &           $257$ &           $-350$ &           $-963$ &          $-6739$ &          $-11124$ &           $-11704$ &           $-11737$ &    $-11741$ \\
                                                                                        & Linear     &              -- &              -- &              -- &               -- &               -- &               -- &                -- &                 -- &                 -- &     $14082$ \\
                                                                                        & Mean Pred. &              -- &              -- &              -- &               -- &               -- &               -- &                -- &                 -- &                 -- &     $18091$ \\
      \hline \multirow{6}{*}{\shortstack[c]{\texttt{power} \\ $n=8132$ \\ $d=4$}}       &            & $0.9\textup{s}$ & $2.1\textup{s}$ & $3.2\textup{s}$ &  $4.3\textup{s}$ &  $6.9\textup{s}$ & $16.1\textup{s}$ &  $58.5\textup{s}$ &  $182.8\textup{s}$ &                 -- &       final \\
                                                                                        & IterGP     &          $-180$ &          $-178$ &          $-194$ &           $-242$ &           $-671$ &          $-1336$ &           $-1409$ &            $-1418$ &                 -- &     $-1448$ \\
                                                                                        & SGPR       &           $220$ &           $107$ &           $-34$ &           $-132$ &           $-263$ &           $-443$ &            $-861$ &            $-1599$ &                 -- &     $-1677$ \\
                                                                                        & GPR        &          $-138$ &          $-145$ &          $-153$ &           $-160$ &           $-166$ &           $-188$ &           $-1356$ &            $-1677$ &                 -- &     $-1677$ \\
                                                                                        & Linear     &              -- &              -- &              -- &               -- &               -- &               -- &                -- &                 -- &                 -- &       $771$ \\
                                                                                        & Mean Pred. &              -- &              -- &              -- &               -- &               -- &               -- &                -- &                 -- &                 -- &     $11539$ \\
     \hline \multirow{6}{*}{\shortstack[c]{\texttt{protein} \\ $n=38870$ \\ $d=9$}}     &            & $0.8\textup{s}$ & $1.6\textup{s}$ & $2.7\textup{s}$ &  $7.1\textup{s}$ & $17.5\textup{s}$ & $54.3\textup{s}$ & $159.7\textup{s}$ &  $376.5\textup{s}$ & $1477.2\textup{s}$ &       final \\
                                                                                        & IterGP     &         $40348$ &         $39982$ &         $39603$ &          $38010$ &          $32803$ &          $32588$ &           $32569$ &            $32546$ &            $32535$ &     $32577$ \\
                                                                                        & SGPR       &         $49759$ &         $48093$ &         $47527$ &          $45613$ &          $44711$ &          $43015$ &           $41832$ &            $40253$ &            $37224$ &     $35016$ \\
                                                                                        & Linear     &              -- &              -- &              -- &               -- &               -- &               -- &                -- &                 -- &                 -- &     $48669$ \\
                                                                                        & Mean Pred. &              -- &              -- &              -- &               -- &               -- &               -- &                -- &                 -- &                 -- &     $55154$ \\
   \hline \multirow{6}{*}{\shortstack[c]{\texttt{skillcraft} \\ $n=2837$ \\ $d=19$}}    &            & $0.4\textup{s}$ & $0.7\textup{s}$ & $1.1\textup{s}$ &  $1.6\textup{s}$ &  $2.4\textup{s}$ &  $9.5\textup{s}$ &  $22.4\textup{s}$ &                 -- &                 -- &       final \\
                                                                                        & IterGP     &          $3395$ &          $3170$ &          $2938$ &           $2837$ &           $2809$ &           $2807$ &            $2804$ &                 -- &                 -- &      $2795$ \\
                                                                                        & SGPR       &          $4026$ &          $4026$ &          $4025$ &           $4025$ &           $4025$ &           $2798$ &            $2797$ &                 -- &                 -- &      $2797$ \\
                                                                                        & GPR        &          $3503$ &          $3341$ &          $3221$ &           $3091$ &           $2874$ &           $2804$ &            $2799$ &                 -- &                 -- &      $2791$ \\
                                                                                        & Linear     &              -- &              -- &              -- &               -- &               -- &               -- &                -- &                 -- &                 -- &      $2792$ \\
                                                                                        & Mean Pred. &              -- &              -- &              -- &               -- &               -- &               -- &                -- &                 -- &                 -- &      $4026$ \\
  \hline \multirow{6}{*}{\shortstack[c]{\texttt{tamielectric} \\ $n=38913$ \\ $d=3$}}   &            & $0.3\textup{s}$ & $0.7\textup{s}$ & $1.2\textup{s}$ &  $2.7\textup{s}$ &  $5.7\textup{s}$ & $16.4\textup{s}$ &  $52.2\textup{s}$ &  $209.7\textup{s}$ & $1722.0\textup{s}$ &       final \\
                                                                                        & IterGP     &         $58526$ &         $57015$ &         $55447$ &          $55289$ &          $55237$ &          $55189$ &           $55032$ &            $54337$ &            $47666$ &     $55236$ \\
                                                                                        & SGPR       &         $55216$ &         $55215$ &         $55215$ &          $57478$ &          $55215$ &          $55215$ &           $55215$ &            $55375$ &            $55215$ &     $55215$ \\
                                                                                        & Linear     &              -- &              -- &              -- &               -- &               -- &               -- &                -- &                 -- &                 -- &     $55214$ \\
                                                                                        & Mean Pred. &              -- &              -- &              -- &               -- &               -- &               -- &                -- &                 -- &                 -- &     $55215$ \\
\hline
\end{tabular}

%% file: tables/rmse.tex
\begin{tabular}{clrrrrrrrrrr}
\hline
                                        Dataset                                         & Method     &    $m=\num{10}$ &      $\num{20}$ &      $\num{50}$ &      $\num{100}$ &      $\num{200}$ &      $\num{500}$ &      $\num{1000}$ &       $\num{2000}$ &       $\num{5000}$ &     final $M$ \\
\hline
      \hline \multirow{6}{*}{\shortstack[c]{\texttt{bike} \\ $n=14772$ \\ $d=17$}}      &            & $1.3\textup{s}$ & $2.4\textup{s}$ & $5.3\textup{s}$ &  $7.2\textup{s}$ & $11.2\textup{s}$ & $25.0\textup{s}$ &  $56.6\textup{s}$ &  $180.0\textup{s}$ &  $955.9\textup{s}$ &         final \\
                                                                                        & IterGP     &         $0.207$ &         $0.166$ &         $0.068$ &          $0.051$ &          $0.046$ &          $0.045$ &           $0.043$ &            $0.048$ &            $0.049$ &       $0.108$ \\
                                                                                        & SGPR       &         $0.191$ &         $0.143$ &         $0.059$ &          $0.055$ &          $0.049$ &          $0.032$ &           $0.039$ &            $0.047$ &            $0.084$ &       $0.048$ \\
                                                                                        & GPR        &         $0.238$ &         $0.236$ &         $0.232$ &          $0.230$ &          $0.224$ &          $0.205$ &           $0.160$ &            $0.061$ &            $0.099$ &       $0.064$ \\
                                                                                        & Linear     &              -- &              -- &              -- &               -- &               -- &               -- &                -- &                 -- &                 -- &       $0.517$ \\
                                                                                        & Mean Pred. &              -- &              -- &              -- &               -- &               -- &               -- &                -- &                 -- &                 -- &       $0.998$ \\
   \hline \multirow{6}{*}{\shortstack[c]{\texttt{elevators} \\ $n=14109$ \\ $d=18$}}    &            & $1.0\textup{s}$ & $3.4\textup{s}$ & $5.1\textup{s}$ &  $7.1\textup{s}$ & $13.0\textup{s}$ & $28.7\textup{s}$ &  $62.7\textup{s}$ &  $166.8\textup{s}$ &  $770.8\textup{s}$ &         final \\
                                                                                        & IterGP     &         $0.500$ &         $0.367$ &         $0.359$ &          $0.357$ &          $0.357$ &          $0.356$ &           $0.356$ &            $0.356$ &            $0.355$ &       $0.545$ \\
                                                                                        & SGPR       &         $0.443$ &         $0.403$ &         $0.398$ &          $0.379$ &          $0.366$ &          $0.358$ &           $0.357$ &            $0.357$ &            $0.356$ &       $0.357$ \\
                                                                                        & GPR        &         $0.459$ &         $0.457$ &         $0.456$ &          $0.454$ &          $0.450$ &          $0.437$ &           $0.409$ &            $0.361$ &            $0.355$ &       $0.356$ \\
                                                                                        & Linear     &              -- &              -- &              -- &               -- &               -- &               -- &                -- &                 -- &                 -- & $\num{3e+09}$ \\
                                                                                        & Mean Pred. &              -- &              -- &              -- &               -- &               -- &               -- &                -- &                 -- &                 -- &       $0.990$ \\
  \hline \multirow{6}{*}{\shortstack[c]{\texttt{keggdirected} \\ $n=41502$ \\ $d=20$}}  &            & $1.6\textup{s}$ & $3.3\textup{s}$ & $5.0\textup{s}$ &  $8.1\textup{s}$ & $15.5\textup{s}$ & $63.6\textup{s}$ & $278.2\textup{s}$ & $1023.8\textup{s}$ & $4854.8\textup{s}$ &         final \\
                                                                                        & IterGP     &         $0.136$ &         $0.135$ &         $0.135$ &          $0.135$ &          $0.133$ &          $0.126$ &           $0.261$ &            $0.145$ &            $0.103$ &       $0.094$ \\
                                                                                        & SGPR       &         $0.146$ &         $0.118$ &         $0.103$ &          $0.098$ &          $0.094$ &          $0.092$ &           $0.092$ &            $0.091$ &            $0.091$ &       $0.091$ \\
                                                                                        & Linear     &              -- &              -- &              -- &               -- &               -- &               -- &                -- &                 -- &                 -- & $\num{6e+13}$ \\
                                                                                        & Mean Pred. &              -- &              -- &              -- &               -- &               -- &               -- &                -- &                 -- &                 -- &       $0.999$ \\
 \hline \multirow{6}{*}{\shortstack[c]{\texttt{keggundirected} \\ $n=54066$ \\ $d=27$}} &            & $2.0\textup{s}$ & $4.3\textup{s}$ & $7.4\textup{s}$ & $12.6\textup{s}$ & $21.1\textup{s}$ & $88.5\textup{s}$ & $485.6\textup{s}$ & $1498.5\textup{s}$ & $8298.0\textup{s}$ &         final \\
                                                                                        & IterGP     &         $0.153$ &         $0.153$ &         $0.153$ &          $0.152$ &          $0.151$ &          $0.145$ &           $0.124$ &            $0.122$ &            $0.123$ &       $0.123$ \\
                                                                                        & SGPR       &         $0.147$ &         $0.130$ &         $0.126$ &          $0.124$ &          $0.122$ &          $0.120$ &           $0.119$ &            $0.118$ &            $0.118$ &       $0.118$ \\
                                                                                        & Linear     &              -- &              -- &              -- &               -- &               -- &               -- &                -- &                 -- &                 -- &       $0.168$ \\
                                                                                        & Mean Pred. &              -- &              -- &              -- &               -- &               -- &               -- &                -- &                 -- &                 -- &       $1.000$ \\
     \hline \multirow{6}{*}{\shortstack[c]{\texttt{kin40k} \\ $n=34000$ \\ $d=8$}}      &            & $0.6\textup{s}$ & $1.3\textup{s}$ & $2.7\textup{s}$ &  $4.5\textup{s}$ &  $8.7\textup{s}$ & $17.7\textup{s}$ &  $44.5\textup{s}$ &  $149.3\textup{s}$ &  $680.0\textup{s}$ &         final \\
                                                                                        & IterGP     &         $0.089$ &         $0.089$ &         $0.089$ &          $0.089$ &          $0.088$ &          $0.088$ &           $0.085$ &            $0.075$ &            $0.074$ &       $0.071$ \\
                                                                                        & SGPR       &         $0.899$ &         $0.810$ &         $0.655$ &          $0.545$ &          $0.425$ &          $0.303$ &           $0.223$ &            $0.163$ &            $0.109$ &       $0.085$ \\
                                                                                        & Linear     &              -- &              -- &              -- &               -- &               -- &               -- &                -- &                 -- &                 -- &       $0.998$ \\
                                                                                        & Mean Pred. &              -- &              -- &              -- &               -- &               -- &               -- &                -- &                 -- &                 -- &       $0.998$ \\
      \hline \multirow{6}{*}{\shortstack[c]{\texttt{kin8nm} \\ $n=6963$ \\ $d=8$}}      &            & $0.6\textup{s}$ & $1.7\textup{s}$ & $3.4\textup{s}$ &  $4.2\textup{s}$ &  $5.8\textup{s}$ &  $8.9\textup{s}$ &  $16.1\textup{s}$ &   $38.3\textup{s}$ &                 -- &         final \\
                                                                                        & IterGP     &         $0.274$ &         $0.259$ &         $0.256$ &          $0.256$ &          $0.256$ &          $0.256$ &           $0.256$ &            $0.256$ &                 -- &       $0.256$ \\
                                                                                        & SGPR       &         $0.707$ &         $0.646$ &         $0.497$ &          $0.466$ &          $0.368$ &          $0.320$ &           $0.285$ &            $0.263$ &                 -- &       $0.256$ \\
                                                                                        & GPR        &         $0.263$ &         $0.263$ &         $0.262$ &          $0.261$ &          $0.260$ &          $0.259$ &           $0.256$ &            $0.256$ &                 -- &       $0.256$ \\
                                                                                        & Linear     &              -- &              -- &              -- &               -- &               -- &               -- &                -- &                 -- &                 -- &       $0.763$ \\
                                                                                        & Mean Pred. &              -- &              -- &              -- &               -- &               -- &               -- &                -- &                 -- &                 -- &       $0.994$ \\
     \hline \multirow{6}{*}{\shortstack[c]{\texttt{naval} \\ $n=10143$ \\ $d=14$}}      &            & $1.9\textup{s}$ & $2.5\textup{s}$ & $4.3\textup{s}$ &  $6.3\textup{s}$ &  $9.0\textup{s}$ & $13.1\textup{s}$ &  $28.6\textup{s}$ &   $78.6\textup{s}$ &                 -- &         final \\
                                                                                        & IterGP     &        $23.910$ &        $32.218$ &        $54.122$ &         $78.432$ &        $111.991$ &        $163.057$ &     $\num{1e+06}$ &      $\num{5e+06}$ &                 -- & $\num{5e+06}$ \\
                                                                                        & SGPR       &         $0.163$ &         $0.061$ &         $0.002$ &          $0.001$ &          $0.001$ &          $0.001$ &           $0.001$ &            $0.001$ &                 -- &       $0.001$ \\
                                                                                        & GPR        &         $0.001$ &         $0.001$ &         $0.001$ &          $0.001$ &          $0.001$ &          $0.001$ &           $0.001$ &            $0.001$ &                 -- &       $0.001$ \\
                                                                                        & Linear     &              -- &              -- &              -- &               -- &               -- &               -- &                -- &                 -- &                 -- &       $0.403$ \\
                                                                                        & Mean Pred. &              -- &              -- &              -- &               -- &               -- &               -- &                -- &                 -- &                 -- &       $1.008$ \\
      \hline \multirow{6}{*}{\shortstack[c]{\texttt{pol} \\ $n=12750$ \\ $d=26$}}       &            & $1.0\textup{s}$ & $2.5\textup{s}$ & $3.8\textup{s}$ &  $8.0\textup{s}$ & $12.2\textup{s}$ & $92.4\textup{s}$ & $310.3\textup{s}$ & $1660.5\textup{s}$ & $7976.3\textup{s}$ &         final \\
                                                                                        & IterGP     &         $0.257$ &         $0.255$ &         $0.253$ &          $0.248$ &          $0.242$ &          $0.178$ &           $0.120$ &            $0.113$ &            $0.101$ &       $0.101$ \\
                                                                                        & SGPR       &         $0.602$ &         $0.390$ &         $0.390$ &          $0.352$ &          $0.295$ &          $0.201$ &           $0.152$ &            $0.116$ &            $0.096$ &       $0.099$ \\
                                                                                        & GPR        &         $0.251$ &         $0.250$ &         $0.249$ &          $0.247$ &          $0.245$ &          $0.200$ &           $0.112$ &            $0.099$ &            $0.099$ &       $0.099$ \\
                                                                                        & Linear     &              -- &              -- &              -- &               -- &               -- &               -- &                -- &                 -- &                 -- &       $0.725$ \\
                                                                                        & Mean Pred. &              -- &              -- &              -- &               -- &               -- &               -- &                -- &                 -- &                 -- &       $0.997$ \\
      \hline \multirow{6}{*}{\shortstack[c]{\texttt{power} \\ $n=8132$ \\ $d=4$}}       &            & $0.9\textup{s}$ & $2.1\textup{s}$ & $3.2\textup{s}$ &  $4.3\textup{s}$ &  $6.9\textup{s}$ & $16.1\textup{s}$ &  $58.5\textup{s}$ &  $182.8\textup{s}$ &                 -- &         final \\
                                                                                        & IterGP     &         $0.232$ &         $0.232$ &         $0.231$ &          $0.224$ &          $0.203$ &          $0.176$ &           $0.175$ &            $0.175$ &                 -- &       $0.175$ \\
                                                                                        & SGPR       &         $0.253$ &         $0.249$ &         $0.243$ &          $0.238$ &          $0.232$ &          $0.225$ &           $0.198$ &            $0.176$ &                 -- &       $0.174$ \\
                                                                                        & GPR        &         $0.231$ &         $0.231$ &         $0.231$ &          $0.230$ &          $0.230$ &          $0.228$ &           $0.188$ &            $0.174$ &                 -- &       $0.174$ \\
                                                                                        & Linear     &              -- &              -- &              -- &               -- &               -- &               -- &                -- &                 -- &                 -- &       $0.272$ \\
                                                                                        & Mean Pred. &              -- &              -- &              -- &               -- &               -- &               -- &                -- &                 -- &                 -- &       $0.996$ \\
     \hline \multirow{6}{*}{\shortstack[c]{\texttt{protein} \\ $n=38870$ \\ $d=9$}}     &            & $0.8\textup{s}$ & $1.6\textup{s}$ & $2.7\textup{s}$ &  $7.1\textup{s}$ & $17.5\textup{s}$ & $54.3\textup{s}$ & $159.7\textup{s}$ &  $376.5\textup{s}$ & $1477.2\textup{s}$ &         final \\
                                                                                        & IterGP     &         $0.625$ &         $0.617$ &         $0.607$ &          $0.565$ &          $0.534$ &          $0.530$ &           $0.530$ &            $0.530$ &            $0.530$ &       $0.530$ \\
                                                                                        & SGPR       &         $0.874$ &         $0.833$ &         $0.819$ &          $0.774$ &          $0.756$ &          $0.721$ &           $0.686$ &            $0.652$ &            $0.589$ &       $0.554$ \\
                                                                                        & Linear     &              -- &              -- &              -- &               -- &               -- &               -- &                -- &                 -- &                 -- &       $0.847$ \\
                                                                                        & Mean Pred. &              -- &              -- &              -- &               -- &               -- &               -- &                -- &                 -- &                 -- &       $1.000$ \\
   \hline \multirow{6}{*}{\shortstack[c]{\texttt{skillcraft} \\ $n=2837$ \\ $d=19$}}    &            & $0.4\textup{s}$ & $0.7\textup{s}$ & $1.1\textup{s}$ &  $1.6\textup{s}$ &  $2.4\textup{s}$ &  $9.5\textup{s}$ &  $22.4\textup{s}$ &                 -- &                 -- &         final \\
                                                                                        & IterGP     &         $0.790$ &         $0.731$ &         $0.696$ &          $0.676$ &          $0.669$ &          $0.668$ &           $0.669$ &                 -- &                 -- &       $0.669$ \\
                                                                                        & SGPR       &         $1.030$ &         $1.030$ &         $1.030$ &          $1.030$ &          $1.030$ &          $0.668$ &           $0.669$ &                 -- &                 -- &       $0.669$ \\
                                                                                        & GPR        &         $0.822$ &         $0.783$ &         $0.737$ &          $0.715$ &          $0.680$ &          $0.668$ &           $0.668$ &                 -- &                 -- &       $0.670$ \\
                                                                                        & Linear     &              -- &              -- &              -- &               -- &               -- &               -- &                -- &                 -- &                 -- &       $0.680$ \\
                                                                                        & Mean Pred. &              -- &              -- &              -- &               -- &               -- &               -- &                -- &                 -- &                 -- &       $1.030$ \\
  \hline \multirow{6}{*}{\shortstack[c]{\texttt{tamielectric} \\ $n=38913$ \\ $d=3$}}   &            & $0.3\textup{s}$ & $0.7\textup{s}$ & $1.2\textup{s}$ &  $2.7\textup{s}$ &  $5.7\textup{s}$ & $16.4\textup{s}$ &  $52.2\textup{s}$ &  $209.7\textup{s}$ & $1722.0\textup{s}$ &         final \\
                                                                                        & IterGP     &         $1.002$ &         $1.002$ &         $1.002$ &          $1.002$ &          $1.002$ &          $1.002$ &           $1.000$ &            $0.993$ &            $0.930$ &       $1.002$ \\
                                                                                        & SGPR       &         $1.002$ &         $1.002$ &         $1.002$ &          $1.002$ &          $1.002$ &          $1.002$ &           $1.002$ &            $1.002$ &            $1.002$ &       $1.002$ \\
                                                                                        & Linear     &              -- &              -- &              -- &               -- &               -- &               -- &                -- &                 -- &                 -- &       $1.002$ \\
                                                                                        & Mean Pred. &              -- &              -- &              -- &               -- &               -- &               -- &                -- &                 -- &                 -- &       $1.002$ \\
\hline
\end{tabular}

%% file: tables/nlpd.tex
\begin{tabular}{clrrrrrrrrrr}
\hline
                                        Dataset                                         & Method     &    $m=\num{10}$ &      $\num{20}$ &      $\num{50}$ &      $\num{100}$ &      $\num{200}$ &      $\num{500}$ &      $\num{1000}$ &       $\num{2000}$ &       $\num{5000}$ &     final $M$ \\
\hline
      \hline \multirow{6}{*}{\shortstack[c]{\texttt{bike} \\ $n=14772$ \\ $d=17$}}      &            & $1.3\textup{s}$ & $2.4\textup{s}$ & $5.3\textup{s}$ &  $7.2\textup{s}$ & $11.2\textup{s}$ & $25.0\textup{s}$ &  $56.6\textup{s}$ &  $180.0\textup{s}$ &  $955.9\textup{s}$ &         final \\
                                                                                        & IterGP     &        $-0.208$ &        $-0.560$ &        $-1.414$ &         $-1.569$ &         $-1.624$ &         $-1.682$ &          $-1.814$ &           $-1.990$ &           $-1.714$ &     $267.942$ \\
                                                                                        & SGPR       &        $-0.264$ &        $-0.551$ &        $-1.418$ &         $-1.489$ &         $-1.584$ &         $-2.045$ &          $-2.421$ &           $-2.778$ &           $-3.797$ &      $-4.052$ \\
                                                                                        & GPR        &         $0.030$ &         $0.017$ &        $-0.016$ &         $-0.038$ &         $-0.084$ &         $-0.245$ &          $-0.612$ &           $-1.930$ &           $-4.035$ &      $-4.111$ \\
                                                                                        & Linear     &              -- &              -- &              -- &               -- &               -- &               -- &                -- &                 -- &                 -- &       $0.759$ \\
                                                                                        & Mean Pred. &              -- &              -- &              -- &               -- &               -- &               -- &                -- &                 -- &                 -- &       $1.417$ \\
   \hline \multirow{6}{*}{\shortstack[c]{\texttt{elevators} \\ $n=14109$ \\ $d=18$}}    &            & $1.0\textup{s}$ & $3.4\textup{s}$ & $5.1\textup{s}$ &  $7.1\textup{s}$ & $13.0\textup{s}$ & $28.7\textup{s}$ &  $62.7\textup{s}$ &  $166.8\textup{s}$ &  $770.8\textup{s}$ &         final \\
                                                                                        & IterGP     &         $0.643$ &         $0.407$ &         $0.392$ &          $0.391$ &          $0.392$ &         $66.141$ &     $\num{4e+03}$ &            $0.411$ &      $\num{5e+03}$ & $\num{2e+05}$ \\
                                                                                        & SGPR       &         $0.600$ &         $0.508$ &         $0.490$ &          $0.443$ &          $0.412$ &          $0.391$ &           $0.387$ &            $0.387$ &            $0.386$ &       $0.386$ \\
                                                                                        & GPR        &         $0.711$ &         $0.704$ &         $0.700$ &          $0.695$ &          $0.679$ &          $0.637$ &           $0.545$ &            $0.396$ &            $0.384$ &       $0.384$ \\
                                                                                        & Linear     &              -- &              -- &              -- &               -- &               -- &               -- &                -- &                 -- &                 -- & $\num{9e+19}$ \\
                                                                                        & Mean Pred. &              -- &              -- &              -- &               -- &               -- &               -- &                -- &                 -- &                 -- &       $1.410$ \\
  \hline \multirow{6}{*}{\shortstack[c]{\texttt{keggdirected} \\ $n=41502$ \\ $d=20$}}  &            & $1.6\textup{s}$ & $3.3\textup{s}$ & $5.0\textup{s}$ &  $8.1\textup{s}$ & $15.5\textup{s}$ & $63.6\textup{s}$ & $278.2\textup{s}$ & $1023.8\textup{s}$ & $4854.8\textup{s}$ &         final \\
                                                                                        & IterGP     &        $-0.658$ &        $-0.661$ &        $-0.664$ &         $-0.670$ &         $-0.684$ &         $-0.774$ &          $11.285$ &            $2.354$ &          $377.359$ &      $-1.029$ \\
                                                                                        & SGPR       &        $-0.536$ &        $-0.762$ &        $-0.883$ &         $-0.936$ &         $-0.992$ &         $-1.020$ &          $-1.038$ &           $-1.045$ &           $-1.050$ &      $-1.050$ \\
                                                                                        & Linear     &              -- &              -- &              -- &               -- &               -- &               -- &                -- &                 -- &                 -- & $\num{1e+29}$ \\
                                                                                        & Mean Pred. &              -- &              -- &              -- &               -- &               -- &               -- &                -- &                 -- &                 -- &       $1.418$ \\
 \hline \multirow{6}{*}{\shortstack[c]{\texttt{keggundirected} \\ $n=54066$ \\ $d=27$}} &            & $2.0\textup{s}$ & $4.3\textup{s}$ & $7.4\textup{s}$ & $12.6\textup{s}$ & $21.1\textup{s}$ & $88.5\textup{s}$ & $485.6\textup{s}$ & $1498.5\textup{s}$ & $8298.0\textup{s}$ &         final \\
                                                                                        & IterGP     &        $-0.255$ &        $-0.261$ &        $-0.268$ &         $-0.281$ &         $-0.302$ &         $-0.465$ &          $-0.712$ &          $826.213$ &           $-0.705$ &      $40.311$ \\
                                                                                        & SGPR       &        $-0.573$ &        $-0.640$ &        $-0.653$ &         $-0.673$ &         $-0.681$ &         $-0.700$ &          $-0.711$ &           $-0.716$ &           $-0.717$ &      $-0.717$ \\
                                                                                        & Linear     &              -- &              -- &              -- &               -- &               -- &               -- &                -- &                 -- &                 -- &      $-0.366$ \\
                                                                                        & Mean Pred. &              -- &              -- &              -- &               -- &               -- &               -- &                -- &                 -- &                 -- &       $1.419$ \\
     \hline \multirow{6}{*}{\shortstack[c]{\texttt{kin40k} \\ $n=34000$ \\ $d=8$}}      &            & $0.6\textup{s}$ & $1.3\textup{s}$ & $2.7\textup{s}$ &  $4.5\textup{s}$ &  $8.7\textup{s}$ & $17.7\textup{s}$ &  $44.5\textup{s}$ &  $149.3\textup{s}$ &  $680.0\textup{s}$ &         final \\
                                                                                        & IterGP     &        $-0.330$ &        $-0.334$ &        $-0.342$ &         $-0.352$ &         $-0.376$ &         $-0.427$ &          $-0.580$ &           $-1.123$ &           $-1.166$ &      $-1.188$ \\
                                                                                        & SGPR       &         $1.313$ &         $1.207$ &         $0.995$ &          $0.814$ &          $0.567$ &          $0.226$ &          $-0.079$ &           $-0.400$ &           $-0.818$ &      $-1.092$ \\
                                                                                        & Linear     &              -- &              -- &              -- &               -- &               -- &               -- &                -- &                 -- &                 -- &       $1.417$ \\
                                                                                        & Mean Pred. &              -- &              -- &              -- &               -- &               -- &               -- &                -- &                 -- &                 -- &       $1.417$ \\
      \hline \multirow{6}{*}{\shortstack[c]{\texttt{kin8nm} \\ $n=6963$ \\ $d=8$}}      &            & $0.6\textup{s}$ & $1.7\textup{s}$ & $3.4\textup{s}$ &  $4.2\textup{s}$ &  $5.8\textup{s}$ &  $8.9\textup{s}$ &  $16.1\textup{s}$ &   $38.3\textup{s}$ &                 -- &         final \\
                                                                                        & IterGP     &         $0.210$ &         $0.078$ &         $0.047$ &          $0.047$ &          $0.047$ &          $0.047$ &           $0.046$ &            $0.047$ &                 -- &       $0.047$ \\
                                                                                        & SGPR       &         $1.071$ &         $0.980$ &         $0.713$ &          $0.646$ &          $0.420$ &          $0.277$ &           $0.156$ &            $0.075$ &                 -- &       $0.048$ \\
                                                                                        & GPR        &         $0.134$ &         $0.127$ &         $0.116$ &          $0.111$ &          $0.101$ &          $0.081$ &           $0.046$ &            $0.047$ &                 -- &       $0.047$ \\
                                                                                        & Linear     &              -- &              -- &              -- &               -- &               -- &               -- &                -- &                 -- &                 -- &       $1.148$ \\
                                                                                        & Mean Pred. &              -- &              -- &              -- &               -- &               -- &               -- &                -- &                 -- &                 -- &       $1.413$ \\
     \hline \multirow{6}{*}{\shortstack[c]{\texttt{naval} \\ $n=10143$ \\ $d=14$}}      &            & $1.9\textup{s}$ & $2.5\textup{s}$ & $4.3\textup{s}$ &  $6.3\textup{s}$ &  $9.0\textup{s}$ & $13.1\textup{s}$ &  $28.6\textup{s}$ &   $78.6\textup{s}$ &                 -- &         final \\
                                                                                        & IterGP     &   $\num{2e+10}$ &   $\num{3e+10}$ &   $\num{5e+10}$ &    $\num{8e+10}$ &    $\num{1e+11}$ &    $\num{2e+11}$ &     $\num{2e+19}$ &      $\num{6e+19}$ &                 -- & $\num{6e+19}$ \\
                                                                                        & SGPR       &        $-0.846$ &        $-2.237$ &        $-4.495$ &         $-4.754$ &         $-4.770$ &         $-4.770$ &          $-4.771$ &           $-4.770$ &                 -- &      $-4.771$ \\
                                                                                        & GPR        &        $-4.731$ &        $-4.732$ &        $-4.734$ &         $-4.736$ &         $-4.739$ &         $-4.743$ &          $-4.760$ &           $-4.773$ &                 -- &      $-4.773$ \\
                                                                                        & Linear     &              -- &              -- &              -- &               -- &               -- &               -- &                -- &                 -- &                 -- &       $0.510$ \\
                                                                                        & Mean Pred. &              -- &              -- &              -- &               -- &               -- &               -- &                -- &                 -- &                 -- &       $1.427$ \\
      \hline \multirow{6}{*}{\shortstack[c]{\texttt{pol} \\ $n=12750$ \\ $d=26$}}       &            & $1.0\textup{s}$ & $2.5\textup{s}$ & $3.8\textup{s}$ &  $8.0\textup{s}$ & $12.2\textup{s}$ & $92.4\textup{s}$ & $310.3\textup{s}$ & $1660.5\textup{s}$ & $7976.3\textup{s}$ &         final \\
                                                                                        & IterGP     &         $0.098$ &         $0.056$ &         $0.021$ &         $-0.092$ &         $-0.206$ &         $-0.837$ &          $-1.076$ &           $-1.111$ &           $-1.146$ &      $-1.149$ \\
                                                                                        & SGPR       &         $0.881$ &         $0.484$ &         $0.484$ &          $0.383$ &          $0.199$ &         $-0.170$ &          $-0.457$ &           $-0.789$ &           $-1.100$ &      $-1.159$ \\
                                                                                        & GPR        &        $-0.146$ &        $-0.157$ &        $-0.166$ &         $-0.196$ &         $-0.226$ &         $-0.796$ &          $-1.120$ &           $-1.158$ &           $-1.159$ &      $-1.158$ \\
                                                                                        & Linear     &              -- &              -- &              -- &               -- &               -- &               -- &                -- &                 -- &                 -- &       $1.097$ \\
                                                                                        & Mean Pred. &              -- &              -- &              -- &               -- &               -- &               -- &                -- &                 -- &                 -- &       $1.416$ \\
      \hline \multirow{6}{*}{\shortstack[c]{\texttt{power} \\ $n=8132$ \\ $d=4$}}       &            & $0.9\textup{s}$ & $2.1\textup{s}$ & $3.2\textup{s}$ &  $4.3\textup{s}$ &  $6.9\textup{s}$ & $16.1\textup{s}$ &  $58.5\textup{s}$ &  $182.8\textup{s}$ &                 -- &         final \\
                                                                                        & IterGP     &        $-0.010$ &        $-0.026$ &        $-0.041$ &         $-0.076$ &         $-0.190$ &         $-0.370$ &          $-0.373$ &           $-0.375$ &                 -- &      $-0.374$ \\
                                                                                        & SGPR       &         $0.046$ &         $0.029$ &         $0.004$ &         $-0.014$ &         $-0.038$ &         $-0.073$ &          $-0.205$ &           $-0.364$ &                 -- &      $-0.378$ \\
                                                                                        & GPR        &        $-0.040$ &        $-0.041$ &        $-0.042$ &         $-0.043$ &         $-0.045$ &         $-0.052$ &          $-0.277$ &           $-0.378$ &                 -- &      $-0.378$ \\
                                                                                        & Linear     &              -- &              -- &              -- &               -- &               -- &               -- &                -- &                 -- &                 -- &       $0.117$ \\
                                                                                        & Mean Pred. &              -- &              -- &              -- &               -- &               -- &               -- &                -- &                 -- &                 -- &       $1.415$ \\
     \hline \multirow{6}{*}{\shortstack[c]{\texttt{protein} \\ $n=38870$ \\ $d=9$}}     &            & $0.8\textup{s}$ & $1.6\textup{s}$ & $2.7\textup{s}$ &  $7.1\textup{s}$ & $17.5\textup{s}$ & $54.3\textup{s}$ & $159.7\textup{s}$ &  $376.5\textup{s}$ & $1477.2\textup{s}$ &         final \\
                                                                                        & IterGP     &         $0.969$ &         $0.959$ &         $0.946$ &          $0.891$ &          $0.746$ &          $0.739$ &           $0.739$ &            $0.738$ &            $0.739$ &       $0.738$ \\
                                                                                        & SGPR       &         $1.282$ &         $1.236$ &         $1.219$ &          $1.163$ &          $1.138$ &          $1.084$ &           $1.044$ &            $0.994$ &            $0.891$ &       $0.822$ \\
                                                                                        & Linear     &              -- &              -- &              -- &               -- &               -- &               -- &                -- &                 -- &                 -- &       $1.253$ \\
                                                                                        & Mean Pred. &              -- &              -- &              -- &               -- &               -- &               -- &                -- &                 -- &                 -- &       $1.419$ \\
   \hline \multirow{6}{*}{\shortstack[c]{\texttt{skillcraft} \\ $n=2837$ \\ $d=19$}}    &            & $0.4\textup{s}$ & $0.7\textup{s}$ & $1.1\textup{s}$ &  $1.6\textup{s}$ &  $2.4\textup{s}$ &  $9.5\textup{s}$ &  $22.4\textup{s}$ &                 -- &                 -- &         final \\
                                                                                        & IterGP     &         $1.185$ &         $1.109$ &         $1.052$ &          $1.025$ &          $1.019$ &          $1.018$ &           $1.018$ &                 -- &                 -- &       $1.019$ \\
                                                                                        & SGPR       &         $1.450$ &         $1.450$ &         $1.450$ &          $1.450$ &          $1.450$ &          $1.016$ &           $1.018$ &                 -- &                 -- &       $1.018$ \\
                                                                                        & GPR        &         $1.215$ &         $1.170$ &         $1.117$ &          $1.085$ &          $1.029$ &          $1.017$ &           $1.017$ &                 -- &                 -- &       $1.019$ \\
                                                                                        & Linear     &              -- &              -- &              -- &               -- &               -- &               -- &                -- &                 -- &                 -- &       $1.037$ \\
                                                                                        & Mean Pred. &              -- &              -- &              -- &               -- &               -- &               -- &                -- &                 -- &                 -- &       $1.450$ \\
  \hline \multirow{6}{*}{\shortstack[c]{\texttt{tamielectric} \\ $n=38913$ \\ $d=3$}}   &            & $0.3\textup{s}$ & $0.7\textup{s}$ & $1.2\textup{s}$ &  $2.7\textup{s}$ &  $5.7\textup{s}$ & $16.4\textup{s}$ &  $52.2\textup{s}$ &  $209.7\textup{s}$ & $1722.0\textup{s}$ &         final \\
                                                                                        & IterGP     &         $1.518$ &         $1.497$ &         $1.469$ &          $1.428$ &          $1.421$ &          $1.421$ &           $1.420$ &            $1.416$ &            $1.381$ &       $1.421$ \\
                                                                                        & SGPR       &         $1.421$ &         $1.421$ &         $1.421$ &          $1.480$ &          $1.421$ &          $1.421$ &           $1.421$ &            $1.425$ &            $1.421$ &       $1.421$ \\
                                                                                        & Linear     &              -- &              -- &              -- &               -- &               -- &               -- &                -- &                 -- &                 -- &       $1.421$ \\
                                                                                        & Mean Pred. &              -- &              -- &              -- &               -- &               -- &               -- &                -- &                 -- &                 -- &       $1.421$ \\
\hline
\end{tabular}